%% file: main.tex
\documentclass[sigconf, nonacm]{acmart}
\usepackage[ruled,vlined]{algorithm2e}

\usepackage{algorithmic}

\newcommand\vldbdoi{10.14778/3626292.3626294}
\newcommand\vldbpages{92 - 105}
\newcommand\vldbvolume{17}
\newcommand\vldbissue{2}
\newcommand\vldbyear{2023}
\newcommand\vldbauthors{\authors}
\newcommand\vldbtitle{\shorttitle} 
\newcommand\vldbavailabilityurl{https://github.com/HazyResearch/evaporate}
\usepackage{balance}

\pdfoutput=1
\usepackage{hyperref}
\usepackage{xcolor}
\usepackage{color}
\usepackage[noabbrev,capitalise,nameinlink]{cleveref}
\hypersetup{colorlinks=true, citecolor=dblue, linkcolor=dblue, urlcolor=dblue}
\usepackage{microtype}
\usepackage{graphicx}
\usepackage{booktabs} % for professional tables
\usepackage{bm}
\usepackage{bbm}
\usepackage{amsmath}  % Define \boldsymbol (in amsbsy too) and align
\usepackage{dsfont}
\usepackage{makecell}
\usepackage{amsthm}
\usepackage{textcomp}
\usepackage{multirow}
\usepackage{mathtools}
\usepackage{geometry}
\usepackage{listings}
\usepackage{boldline}
\usepackage{tabularx}
\usepackage[export]{adjustbox}
\usepackage{mathtools}
\usepackage{graphicx}
\usepackage{xspace}
\usepackage{bbm}

\usepackage{tcolorbox}
\newtcolorbox{mybox}{colback=blue!5!white,colframe=blue!5!black}

\definecolor{mygray}{gray}{0.6}
\definecolor{keywordcolor}{rgb}{0.2,0.2,0.6}
\definecolor{keywordcolor2}{rgb}{0.15,0.46,0.1}
\definecolor{typecolor}{rgb}{0.17,0.56,0.68}
\definecolor{commentcolor}{gray}{0.3}
\definecolor{ratecolor}{rgb}{0.5,0.1,0.1}
\definecolor{stringcolor}{gray}{0.3}

\usepackage{bm}
\usepackage{enumitem}
\usepackage{caption}
\usepackage{subcaption}

\usepackage{listings}
\usepackage{color}
\usepackage{xcolor}

\usepackage[T1]{fontenc}
\usepackage{mdframed} %nice frames

\definecolor{dkgreen}{rgb}{0,0.6,0}
\definecolor{gray}{rgb}{0.5,0.5,0.5}
\definecolor{light-gray}{gray}{0.95} %the shade of grey that stack exchange uses
\definecolor{mauve}{rgb}{0.58,0,0.82}
\definecolor{backcolour}{rgb}{0.95,0.95,0.92}

\newmdenv[linecolor=light-gray,%
backgroundcolor=light-gray,
innerleftmargin=2.8pt,
innerbottommargin=-0.8pt,
leftmargin=0.0pt,
rightmargin=0.0pt,
skipbelow=-2.0pt,
frametitle={}]{codeframe}

\lstdefinestyle{prompts}{  
    commentstyle=\color{dkgreen},
    keywordstyle=\color{magenta},
    moredelim=**[is][\color{mauve}]{@}{@},
    basicstyle=\fontsize{7}{9}\selectfont\ttfamily,
    breakatwhitespace=false,         
    breaklines=true,                 
    captionpos=b,                    
    keepspaces=true,                 
    numbers=none,                    
    numbersep=5pt,                  
    showspaces=false,
    showstringspaces=false,
    showtabs=false,                  
    tabsize=2
}
\definecolor{mycolor}{rgb}{0.122, 0.435, 0.698}

\newcommand*{\ShowNotes}{}

\input{macros.tex}

\newif\ifarxiv

\newcommand\vldbpagestyle{plain} 
\newcommand{\systemname}{\textsc{Evaporate}\xspace}
\newcommand{\systemnamea}{\textsc{Evaporate-Direct}\xspace}
\newcommand{\systemnameb}{\textsc{Evaporate-Code}\xspace}
\newcommand{\systemnamec}{\textsc{Evaporate-Code+}\xspace}

\newcommand{\edit}[1]{\textcolor{black}{#1}}

\begin{document}

% \input{response.tex}
% \newpage

\title{Language Models Enable Simple Systems for Generating Structured Views of Heterogeneous Data Lakes}

%% The "author" command and its associated commands are used to define the authors and their affiliations.

\settopmatter{authorsperrow=4}
\author{Simran Arora}
\affiliation{\institution{Stanford University}}
\email{simarora@stanford.edu}

\author{Brandon Yang}
\affiliation{\institution{Stanford University}}
\email{bcyang@stanford.edu*}

\author{Sabri Eyuboglu}
\affiliation{\institution{Stanford University}}
\email{eyuboglu@stanford.edu*}

\author{Avanika Narayan}
\affiliation{\institution{Stanford University}}
\email{avanikan@stanford.edu}

\author{Andrew Hojel}
\affiliation{\institution{Stanford University}}
\email{ahojel@stanford.edu}

\author{Immanuel Trummer}
\affiliation{\institution{Cornell University}}
\email{itrummer@cornell.edu}

\author{Christopher Ré}
\affiliation{\institution{Stanford University}}
\email{chrismre@stanford.edu}

\begin{abstract}
A long standing goal in the data management community is developing systems that input documents and output queryable tables without user effort. Given the sheer variety of potential documents, state-of-the art systems make simplifying assumptions 
and use domain specific training. 
In this work, we ask whether we can maintain generality by using the in-context learning abilities of large language models (LLMs). 
We propose and evaluate \systemname{}, a prototype system powered by LLMs. We identify two strategies for implementing this system: prompt the LLM to directly extract values from documents or prompt the LLM to synthesize code that performs the extraction. Our evaluations show a cost-quality tradeoff between these two approaches. Code synthesis is cheap, but far less accurate than directly processing each document with the LLM. 
To improve quality while maintaining low cost, we propose an extended implementation, \systemnamec, which achieves better quality than direct extraction. Our insight is to generate many candidate functions and ensemble their extractions using weak supervision. 
\systemnamec outperforms the state-of-the art systems using a \textit{sublinear} pass over the documents with the LLM. This equates to a 110$\times$ reduction in the number of documents the LLM needs to process across our 16 real-world evaluation settings.
\end{abstract}

\maketitle

\pagestyle{\vldbpagestyle}
\begingroup\small\noindent\raggedright\textbf{PVLDB Reference Format:}\\
\vldbauthors. \vldbtitle. PVLDB, \vldbvolume(\vldbissue): \vldbpages, \vldbyear.\\
\href{https://doi.org/\vldbdoi}{doi:\vldbdoi}
\endgroup
\begingroup
\renewcommand\thefootnote{}\footnote{\noindent
* Equal contribution. \newline
This work is licensed under the Creative Commons BY-NC-ND 4.0 International License. Visit \url{https://creativecommons.org/licenses/by-nc-nd/4.0/} to view a copy of this license. For any use beyond those covered by this license, obtain permission by emailing \href{mailto:info@vldb.org}{info@vldb.org}. Copyright is held by the owner/author(s). Publication rights licensed to the VLDB Endowment. \\
\raggedright Proceedings of the VLDB Endowment, Vol. \vldbvolume, No. \vldbissue\ \%
ISSN 2150-8097. \\
\href{https://doi.org/\vldbdoi}{doi:\vldbdoi} \\
}\addtocounter{footnote}{-1}\endgroup

\ifdefempty{\vldbavailabilityurl}{}{
% \vspace{.3cm}
\begingroup\small\noindent\raggedright\textbf{PVLDB Artifact Availability:}\\
The source code, data, and/or other artifacts have been made available at \url{https://github.com/HazyResearch/evaporate}.
\endgroup
}

\section{Introduction}
\input{sections/intro2.tex}

\section{Preliminaries}
\input{sections/problem}

% \input{sections/observations}

% \subsection{}
\label{sec:approach_schema}
\input{sections/framework.tex}

% \vspace{-2mm}
\section{Evaluations}
\label{sec:evaluations}
\input{sections/analysis.tex}

\input{sections/evaluations.tex}

% \vspace{-2mm}
\section{Related Work}
\input{sections/related_works.tex}

% \vspace{-2mm}
\section{Conclusion}
\input{sections/conclusion}

%\clearpage

\bibliographystyle{ACM-Reference-Format}
\bibliography{sample}

\appendix

\input{sections/appendix}

\input{sections/input_output}

\end{document}

%% file: macros.tex
\newtheorem{definition}{Definition}

\newcommand{\argmax}[2]{\textrm{argmax}_{#1}~#2}
\ifdefined\ShowNotes
  \newcommand{\colornote}[3]{{\color{#1}\bf{#2 #3}\normalfont}}
\else
  \newcommand{\colornote}[3]{}
\fi

\definecolor{darkred}{rgb}{0.7,0.1,0.1}
\definecolor{darkgreen}{rgb}{0.1,0.5,0.1}
\definecolor{cyan}{rgb}{0.7,0.0,0.7}
\definecolor{dblue}{rgb}{0.2,0.2,0.8}
\definecolor{maroon}{rgb}{0.76,.13,.28}
\definecolor{burntorange}{rgb}{0.81,.33,0}
\definecolor{royalpurple}{rgb}{0.47,.31,0.66}

\newcommand {\CITE}[1]{\colornote{magenta}{CITE}{}}
\newcommand {\tasks}[1]{data cleaning and integration }
\ifdefined\ShowNotes
  
\else
  
\fi

%% file: sections/intro2.tex
Organizations often seek insights trapped in heterogeneous data lakes (\textit{e.g.} the web, corporate data lakes, and electronic health records) \cite{romero2013edscience, boylen2019bioscience, faghmous2014climatescience}.
In their raw form, these data sources cannot easily support analytical queries. A long standing goal of  the data management community is to develop systems that automatically convert  heterogeneous data lakes into queryable, structured tables ~\cite[inter alia.]{brin1998extraction,
cafarella2007structured, yafooz2013managing, nargesian2019data}. In this work, we investigate whether recent large language models can help address this problem.

We study systems that take as \textbf{input} heterogeneous documents (\textit{e.g.} HTML webpages, PDFs, text) and \textbf{output} a tabular, structured view of the documents. These systems must identify the schema and perform extraction to populate the table.
% \vspace{-2mm}
\begin{mybox}
\textsc{Example 1.} 
Medical researchers frequently use data spanning  electronic health records (EHR), clinical trials, knowledge sources (e.g. PubMed), and FDA reports to understand and monitor patients and treatments \cite{bates2023safety}. Consider the large collection of \textbf{FDA 510(k)} reviews for premarket medical devices, which have been the subject of multiple studies \cite{wu2021medai, zuckerman2011devicerecalls}. Our objective is to output a table that automatically structures the attributes that are distributed in the $\sim$20-page \textbf{PDFs}, for instance the \verb|device classification|, \verb|predicate device code|, and \verb|indications for use|. 
\end{mybox} 
\label{example1}
% \vspace{-2mm}

Systems designed to tackle this problem must balance a three-way tradeoff between \textbf{cost} (data lakes may hold millions of documents), \textbf{quality} (output tables should be able to accurately support an analyst's queries), and \textbf{generality} (different data lakes have different document types and structure). See \cref{sec:definition} for a formal task definition and further discussion of this tradeoff.

\begin{figure*}
    \centering
    \includegraphics[width=\linewidth]{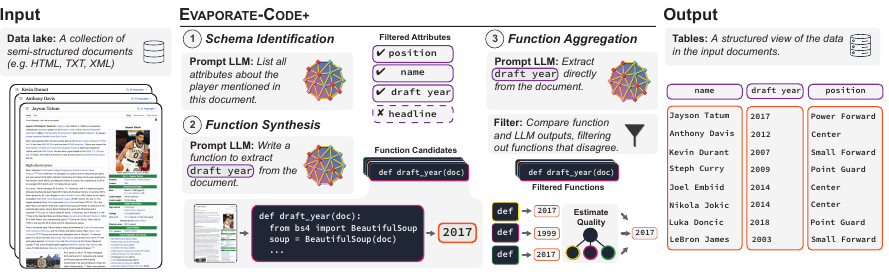}
    \caption[width=\linewidth]{The user provides a collection of documents (e.g. NBA player bios) and \systemname{} outputs a table by identifying attributes and populating columns. \systemname{} avoids running expensive LLM inference on all documents by (1) synthesizing the key attributes from a small sample of documents and (2) synthesizing (e.g. Pythonic) functions that then are reused at scale to process documents. Because function quality is variable, \systemname{} (3) applies an algorithm that generates many candidate functions and ensembles their extractions using weak supervision.}
    \label{fig:main}
    % \vspace{-3mm}
\end{figure*}

% Paragraph Objective: Why do previous approaches fail? 
Given the range of formats, attributes, and domains across documents, prior systems 
rely on simplifying assumptions 
(e.g. handling one document format). The majority of works focus on structuring \textsc{HTML}~\cite{brin1998extraction, etzioni2004knowitall, cafarella2007structured}, assuming the attributes and values are at specific positions in the \textsc{HTML}-DOM~\cite{lockard2019ceres, lockard2020zeroshotceres, deng2022domlm, zhou2022survey}. For unstructured text, current approaches use linguistic tools (e.g., dependency parsers) to introduce structure~\cite{agichtein2000snowball, etzioni2004knowitall, cafarella2007structured, niklaus2018openie} and then apply heuristic rules over the resulting structure to extract information. The documents in Example \ref{example1} highlight the limitations of the prior approaches: they lack (e.g. \textsc{HTML}) structure and, consistent with recent evaluation efforts \cite{zhou2022survey}, we find the SoTA approaches for unstructured text perform poorly on long semi-structured PDFs (See \cite{arora2023evaporate}). Some systems assume there is a human-in-the-loop, labeling data and writing heuristic rules for extraction ~\cite{shin2015incremental,ratner2017snorkel}, while others assume access to annotated training documents from the domain~\cite{lockard2019ceres, lockard2020zeroshotceres, deng2022domlm}. Researchers manually annotated the reports in Example \ref{example1} \cite{wu2021medai}.

In this work, we explore whether we can improve generality by leveraging \textit{large language models} (LLMs). An LLM is a deep learning model that is pretrained on broad data and can be adapted to diverse tasks, from machine translation to data wrangling \cite{brown2020language, narayan2022can}. At inference time, the models take as input a natural language task description termed a \textit{prompt} ~\cite{brown2020language, bommasani2021fm} and generate a natural language response. 
See \cref{sec:llm-background} for more background on LLMs. 

\textbf{\systemname.} (\cref{sec:system}) We present \systemname, a system that uses LLMs to produce structured views of semi-structured data lakes. Our evaluation spans 16 real-world settings from movie and university websites to \textit{FDA 510(k)} reviews ~\cite{hao2011swde, lockard2019ceres, deng2022domlm, klimt2004enron, heller2017enronnyt,wu2021medai,zuckerman2011devicerecalls}.

The user inputs a collection of documents and \systemname automatically identifies the schema and performs extraction to populate the table. Our implementation requires \textit{no customization, training, or human effort} to support the diverse evaluation settings.
We propose two fundamental strategies for implementing this interface, identifying a tradeoff between their cost and quality: 
\begin{enumerate}
    \item \systemnamea (\cref{fig:end2end}) The LLM directly extracts values from documents.
    \item \systemnameb (\cref{fig:function_prompts}) The LLM \textit{synthesizes code} that is then applied to process documents at scale.
    % the values.
\end{enumerate}
\systemnameb is cheap, but underperforms \systemnamea by 24.9\% (13.8 F1 points) averaged across our evaluation settings. We thus seek a new code synthesis approach. We present \systemnamec, which achieves better quality than direct extraction. Our insight is to synthesize many code snippets for extraction and ensemble their outputs using weak supervision.

\textbf{Direct Extraction (\cref{sec:approach-direct}).} Our first implementation, \systemnamea, applies a \textit{single prompt} (included in \cite{arora2023evaporate}) to each document in the input. The prompt instructs the LLM to both identify the schema and extract values. 
Remarkably, we find that in some settings, with a single prompt and no task specific modifications, performance is already competitive with state-of-the-art systems that rely on domain specific assumptions and training. 

However, this implementation is very expensive. LLMs are optimized for interactive, human-in-the-loop applications (\textit{e.g.} ChatGPT) \cite{wu2022ai}, not high-throughput data processing tasks~\cite{sheng2023high}. The number of tokens processed by an LLM in \systemnamea grows \textit{linearly} with the size of the data lake. As of March 2023, applying OpenAI's models to the 55 million Wikipedia articles would cost over \$110k (\verb|gpt-3.5|, \$$0.002/1$k tokens) and \$1.1M (\verb|text-davinci-003|, \$$0.02/1$k tokens) dollars \cite{wiki_deets, openai}. There are \textit{billions} of webpages on the broader Internet \cite{siteefy2023} and the facts change over time. For instance, NBA players are added to Wikipedia, a player's \verb|team| changes after trades, and the \verb|points per game| metric changes after every game. Data processing is a \textit{routine expense} (repeated by multiple data analysts), not a one-time cost ~\cite{shankar2022operationalizing}.

\textbf{Code Synthesis (\cref{sec:approach-code}).} \textit{Can we produce the structured table using a sublinear pass of the LLM over the documents?} We propose 
\systemnameb, which splits the task into two sub-tasks: (1) identify the table schema and (2) extract values. This view allows us to exploit the distinct \textit{redundancies} of each sub-task that occur when running LLM inference on every document:
% \vspace{-2mm}
\begin{enumerate}
\item \textit{Schema Generation.} In order to identify a schema, we only process a small sample of documents with the LLM. This succeeds because there is redundancy in the attributes mentioned across documents. For e.g., in Example \ref{example1}, most reports mention a \verb|predicate device name|.
\item \textit{Function Synthesis.} We prompt the LLM to synthesize (\textit{e.g.} Pythonic) \textit{functions}, that can be applied at scale across the documents. This works because of redundancy in the formatting of attribute-value pairs. For e.g., the FDA 510(k)s use the consistent format ``\verb|Predicate device name: k|".
\end{enumerate}

% \vspace{-1mm}
The number of tokens processed by the LLM in \systemnameb is \textit{fixed} and does not grow with the size of the data lake (as illustrated in \cref{fig:crossinglines}), addressing the cost issues of \systemnamea{}. However, the LLM synthesizes variable quality information extraction functions. The extractions are up to 14 points worse in Pair F1 score than those produced using \systemnamea{}.

\textbf{Code Synthesis + Aggregation.} (\cref{sec:approach-codeplus}) To improve quality while keeping costs low, we propose \systemnamec. Studying the synthesized functions, we observe some only work for a narrow slice of documents, while others exhibit syntactic and logical errors. To reduce variance, we synthesize many candidate functions, then estimate their quality and aggregate their extractions using \textit{weak supervision}. This builds on our work \cite{arora2022ama}, which broadly applies weak supervision to prompting for the first time.

Weak supervision (WS) is a statistical framework for modeling and combining noisy sources with varied coverages without any labeled data~\citep{ratner2017snorkel, varma2019learning}. However, WS is typically applied over \textit{human-generated} functions while our setting consists of \textit{machine-generated} functions. This presents issues when attempting to apply existing WS tools. (1)  WS theoretically assumes all noisy sources are better than random performance (50\% accuracy), yet 40\% of our generated functions are \textit{below 25\%} (\cref{sec:approach}). (2) WS attempts to deploy functions that achieve high quality on narrow slices of data (high precision), and allow the function to \textit{abstain} on data external to the slice (low recall). 
While humans can express when functions should abstain, the machine-generated functions do not contain this logic.
To handle the \textit{open} WS setting, we introduce a new  algorithm for ensembling the functions (\cref{alg:approach}).

We summarize our overall contributions as follows. 
% \vspace{-2mm}
\begin{enumerate}
% [itemsep=0.1pt,topsep=1pt,leftmargin=*]
    \item \textbf{Our system offers new capabilities for the long-studied structured view generation problem.} Existing systems require in-domain training and handle limited document formats (e.g. HTML \cite{cafarella2007navigating, etzioni2004web, lockard2019ceres, deng2022domlm}). \systemname requires no training and succeeds on different document formats (HTML, PDF, TXT) off-the-shelf.
    % , while outperforming the existing work  
    (\cref{sec:analysis}).
    \item \textbf{We study a new tradeoff space between direct extraction and code synthesis for data tasks.} \systemname \textit{asymptotically} reduces the number of tokens that the LLM needs to process to generate the outputs. At 10k documents per evaluation setting, this amounts to a 110x cost reduction. Further, prior works using LLMs for data tasks require users to manually write prompts~\cite{narayan2022can}. \systemname is built with task-agnostic prompts that generalize across settings.
    \item \textbf{We present an algorithm and theoretical analysis for applying weak supervision to open-ended functions and extraction tasks. } Although \systemnameb{} is more efficient,  \systemnamea achieves significantly higher quality. Using our algorithm, \systemnamec{} outperforms \systemnamea, which directly processes every document, by 10.1 F1 points (18\%) (Table \ref{tab:evaporate_vs_fm_openie}).
    \item \textbf{We extensively validate the system on 16 data settings from 5 domains and 3 data formats, and across 4 LLMs.} (1) \systemname outperforms the SoTA \textit{learned} baseline systems by 3.2 F1 points (6\%) when generating tables (both schema generation and extraction) end-to-end, and 6.7 F1 points (10\%) on the extraction step. (2) \systemnamec achieves a 10.1 F1 point increase over \systemnamea{}, using text-davinci-003. 
    (3) Across four unique LLMs we show the relative quality of \systemnamea{} vs. \systemnamec{} remains consistent.
\end{enumerate}

We define the problem in \cref{sec:definition}. We present \systemname{} in \cref{sec:system}, evaluations in \cref{sec:evaluations}, and related works in \cref{sec:related_work}.

%% file: sections/problem.tex
\label{sec:definition}
We first define the problem setting and system desiderata. 

\subsection{Problem Setting}
We study the problem of constructing a structured view (\textit{i.e.} database table) of a set of semi-structured documents (\textit{e.g.} HTML, PDF, TXT). Formally, we define the problem as follows:
\begin{itemize}
    \item \textbf{Input:} User provides a set of $n$ semi-structured documents $D = \{d_1, d_2, ... d_n\}$ (\textit{e.g.} A collection of FDA 510(k) reviews for premarket notification submission for medical devices).
    \item \textbf{Output:} System outputs a table defined by a set of attribute names $A = \{a_1, a_2, ... a_m\}$ (\textit{e.g.} $a_1=$\verb|indications for use|, $a_2$=\verb|classification|) and a set of $n$ extracted records for $R = \{r_1, r_2, ... r_n\}$, one per document, where $r_i$ is an $m$-tuple (\textit{e.g.} $r_1=$ (``fracture", ``x-ray")). 
\end{itemize}
Unlike prior work which proposes systems that rely on manual labeling \cite{shin2015incremental} or manual prompt tuning \cite{narayan2022can, trummer2022codexdb}, we aim to develop \textit{automated} solutions, which require no user effort.

\textit{Measuring System Quality} We compare the generated table $(A, R)$ to a manually curated ``ground-truth" table  $(\hat{A}, \hat{R})$. The coverage of an attribute refers to the fraction of documents that include the attribute and its value. Following prior work, we prioritize attributes with high \textit{coverage}, which tend to be useful for analysis
~\cite{cafarella2007navigating, chu2007relational}. 
We measure agreement between the tables using Pair F1. For additional details on our evaluation setup, see \cref{sec:eval}.

\subsection{System Desiderata} 
\label{sec:desiderata}
Current systems for producing structured views are limited in their generality, cost/flexibility, and quality/usability \cite{cafarella2007navigating, chu2007relational, niklaus2018openie, zhou2022survey}. Here we review the existing systems.

\textbf{Generality.} \textit{The ideal system will generalize across document formats and domains, without manually engineered rules or task-specific training.} This is important because the input documents $D$ could focus on any imaginable topic or use any file format~\cite{zhou2022survey}. 
Existing systems featurize documents by tagging the named entities (NER), dependency parse tree, and part-of-speech (POS), and train a model to predict whether a span of text is a useful fact \cite{kolluru2020openie}.  
Unfortunately, the performance of the parse, NER, and POS tags drastically degrade on semi-structured data (e.g. \textsc{HTML} elements) and  longer sequences of text (i.e. full documents)~\cite{zhou2022survey}. We provide detailed error analysis \cite{arora2023evaporate}.
A specialized class of systems focuses on processing semi-structured web \textsc{HTML} documents by leveraging the HTML DOM tree as features \cite[inter alia.]{etzioni2004knowitall, cafarella2007navigating, bronzi2013extraction, lockard2020zeroshotceres, deng2022domlm}. However, the systems thus do not support other document formats.

\textbf{Cost.} \textit{The ideal system will enable  users to manage a cost-coverage tradeoff, rather than requiring them to extract ``all-or-nothing''. }
% The next key tradeoff is that reducing the cost of current OpenIE systems increases the required human effort.
The existing systems are built to extract \textit{all} possible facts in the documents, without prioritizing important attributes or allowing the user to influence what is extracted \cite{cui2018neuralopenie, zhou2022survey}. 
Processing every line of every document can be expensive. To mitigate this, the user can define the attributes of interest then apply a closed IE system for extraction, however this requires upfront human effort. \textbf{Desiderata}: The ideal system will enable  users to manage a cost-coverage tradeoff, rather than requiring them extract ``all-or-nothing''.

 \textbf{Quality.} \textit{The ideal system will output a table $(A, R)$ with full columns (i.e. high-coverage attributes) and accurate, consistently formatted extractions.} Existing OpenIE systems commonly extract tuples in unnormalized forms directly from documents \cite{cui2018neuralopenie}. This can make the resulting extractions difficult to use for analysis, requiring advanced systems or user-defined post-processing code for resolving subject, objects, and predicates to a canonical form \cite{cafarella2007structured}.

\subsection{Background on Large Language Models}
\label{sec:llm-background}

In this section, we provide background on \textit{large language models} (LLMs), which are central to our work.

\begin{definition}[Large Language Model]
A machine learning model, $\mathcal{F}$, trained on a self-supervised task (\textit{e.g.} next word prediction) over a massive corpus of text~\cite{gao2021pile}. Language models can be used to generate new text based on provided context. For example:
$$
\mathcal{F}(\text{All that glitters }) \rightarrow \text{is not gold.}
$$
\end{definition}

Numerous studies have demonstrated LLMs capable of solving new tasks without updating any model parameters, a phenomenon termed \textit{in-context learning}~\cite{brown2020language,agrawal2022clinicalie,narayan2022can}. 
Specifically, these studies show that when passed an appropriate description of the task, the model often generates text completing the task. 

\begin{definition}[Prompt]
A natural language task-specification used to elicit a particular generation from an LLM. Prompts often include demonstrations of the task.
For example, the prompt below elicits the translation of the word cheese into French: 
$$
\mathcal{F}(
\underbrace{\text{Translate. Eng: hello, Fr: bonjour; Eng: cheese, Fr: }}_{\mathclap{\textbf{Prompt}}}
) \rightarrow \underbrace{ \text{fromage}}_{\mathclap{\textbf{Generation}}}
$$
\end{definition}

Examples of prompts used in this work are provided in \cref{fig:end2end,fig:function_prompts}. All prompts used in the system are provided in \cite{arora2023evaporate}.

%% file: sections/framework.tex
\section{\systemname{}: A Prototype System Powered by Language Models } 
\label{sec:system}

\begin{figure}
    \centering
    \includegraphics[width=0.9\columnwidth]{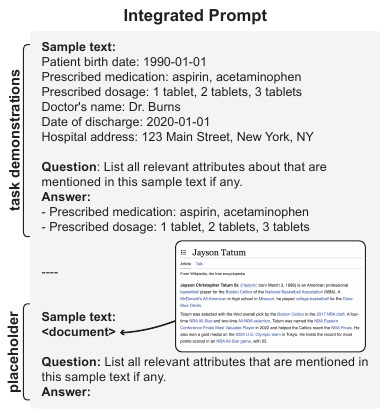}
    \caption[width=\columnwidth]{Prompt for \systemnamea{} structured. The prompt template, which includes placeholders for in-context examples and and the inference example (i.e., data lake documents), is applied to each document in the data lake.}
    \label{fig:end2end}
    % \vspace{-3mm}
\end{figure}

We introduce \systemname{}, a prototype system that uses LLMs to materialize a structured view of a heterogeneous, semi-structured data lake. Compared to prior systems, which rely on manual labeling \cite{shin2015incremental} or tuning prompts to a domain \cite{narayan2022can}, \systemname{} exposes a remarkably \textit{general} interface: the user inputs documents and the system automatically outputs a structured view of those documents, without any domain specific training or prompt customization. 

\textit{Overview.}  We instantiate the \systemname{} interface with three different implementations. We can feed every document to the LLM and prompt it to extract values directly (\textit{direct extraction}, \cref{fig:end2end}), or feed a small sample of documents to the LLM and prompt it to write \textit{code} to do the extraction (\textit{code extraction}, \cref{fig:function_prompts}). In \cref{sec:approach-direct} and \cref{sec:approach-code}, we describe baseline implementations of these two strategies, \systemnamea and \systemnameb. We find that these two implementations tradeoff cost and quality. Then, in \cref{sec:approach-codeplus}, we propose a code extraction implementation that uses weak supervision to improve quality and retain low cost. 

\textit{Prompt Management} \systemname{} applies a set of task-agnostic prompts, which are all provided verbatim in the technical report \cite{arora2023evaporate}.
These tasks are not modified for different tasks. Within the system, the prompts are Python f-strings, with placeholders for inputs chunks of text from the particular dataset being processed. The LLM is prompted with the formatted strings. We use a caching tool that we helped develop called \textsc{Manifest} \cite{orr2022manifest} to store input and completion pairs from the LLM prompting in a local SQLite database, where keys are the prompt-inputs and values are the completions . Therefore, if users repeatedly run the system on the same dataset, they do not incur the LLM inference costs again.

\subsection{\systemnamea}
\label{sec:approach-direct}

In this section, we describe a simple \textit{direct extraction} implementation, \systemnamea that applies a single prompt template to every document. This prompt template, which is included in \cref{fig:end2end}, instructs the LLM to both identify the schema and extract values (see \cite{arora2023evaporate} for the full prompt). It consists of a few in-context examples that are general, \textit{i.e.} are not customized to a particular format, domain, or document. 

Below we discuss how we (1) manage long documents that cannot fit in the LLM's context window, (2) process the LLM's textual outputs, (3) prioritize the most useful attributes according to principles described in prior work \cite{cafarella2007navigating}.

\textit{Managing long documents.} The input to \systemname{} is a file path to raw documents, which can be several pages long. For instance the Medical FDA reports in Example \ref{example1} are $\sim$20 pages long. However, the underlying Transformer architecture of modern LLMs is limited to processing a fixed number of tokens (\textit{e.g.} a few thousand tokens), referred to as the \textit{context window}, during each inference call. \systemname{} therefore splits the raw documents such that each piece is within the context window. Each chunk is inserted into the prompt in turn as shown in \cref{fig:end2end}.

\textit{Processing text outputs.} Language models output open ended text so the last step is to convert this to a usable table. To facilitate this data transformation, we can specify formats in our prompt demonstrations to encourage the LLM to organize the output in a similar structure. For instance, the demonstration in \cref{fig:end2end} specifies a list format with \verb|<attribute>: <value(s)>| per entry. \systemname{} outputs in this format can be de-serialize into a table.

\textit{Prioritizing common attributes.} The list of extracted attributes and values can contain the niche attributes for specific documents, whereas a common database design principle is to capture the high frequency attributes  \cite{cafarella2007structured}. Therefore \systemname{} takes the union of attributes outputted across documents and ranks by frequency to enable prioritizing head attributes.

\textbf{Analysis.} We analyze this \textit{direct extraction} implementation, \systemnamea{}, along the axes of our three desiderata. Results processing the documents with \systemnamea{} are reported in \cref{tab:evaporate_vs_fm_openie} and are discussed in detail in \cref{sec:analysis}. 

Overall, the quality matches or exceeds the baseline systems (described in \cref{sec:analysis}), on 8 of the 16 settings. This is surprising given the simplicity --- i.e. \systemnamea{} uses \textit{one} fixed prompt to process \textit{all 16 settings}. However, fundamental cost limitations impede the real-world deployment of this approach.

However, the high cost of this implementation limits its applicability to large, recurring workloads. The number of tokens processed by the LLM scales linearly with the size of the data lake, $\mathcal{O}(n)$.
Data lakes can contain \textit{billions} of documents \cite{siteefy2023, nargesian2019data}. Further, in most organizations, data processing is not a one time cost. Data lakes are dynamically changing, so \systemnamea would need to be \textit{repeatedly} applied.

\subsection{\systemnameb}
\label{sec:approach-code}

In this section, we present \systemnameb, which significantly reduces cost compared to \systemnamea. Here, we perform schema identification separately from value extraction, which allows us to exploit fundamental differences between the sub-tasks to reduce cost. In schema identification, we find that we only need to process a small sample of documents because attributes are consistent across documents. On the other hand, in order to extract values, we must process every document. However, the ways in which values appear across documents (\textit{i.e.} their relative positions in the document) tend to be consistent, meaning the extraction logic is consistent across documents. 

The two steps of the decomposed implementation are:

\begin{enumerate}
    \item \textbf{Schema synthesis.} (\cref{sec:approach_schema}) We observe that the attribute outputs contain relatively consistent \verb|<attributes>|, even though the \verb|values| differ from document to document. To exploit this redundancy, \systemnamea{} prompts an LLM to analyze a small sample of documents to identify attributes for the output schema For example, given a sample of the Medical Device FDA Reports, the LLM outputs a devices table with attributes like \verb|"510(k) number"|. 
    \item \textbf{Function synthesis} (\cref{sec:approach_extract}). We observe consistencies in how attributes are embedded across documents. E.g., the \verb|510(k) code| in the FDA documents always starts with the letter ``k'' and the \verb|player position| attribute is always in the \textsc{HTML} ``infobox'' element in NBA player Wiki pages. A researcher would likely exploit such redundancies when manually scraping the documents for analysis. In \systemnameb{}, we propose to use the LLM to automatically synthesize a data-lake-specific suite of \textit{functions}, that can then be applied at scale to process many documents.
\end{enumerate}

Next, we provide details for each sub-task.

\label{sec:approach}

% \vspace{-2mm}
\subsubsection{Schema Synthesis }
\label{sec:approach_schema}

 \systemname{} first uses an LLM to identify attributes $A = \{a_1, a_2, ... a_m\}$ for the output schema. 
 
\textit{Generating candidate attributes} Concretely, we sample a set $\tilde{D}$ of $k << n$ documents from $D$. For each, we prompt the LLM to extract the most useful attributes from the document as in \systemnamea{}. Recall this yields a set of attributes ranked by how frequently they were extracted across documents. We retain attributes that are explicitly mentioned in the document to ensure provenance in schema identification.
 
\textit{Re-ranking candidate attributes}  Because \systemname{} now identifies the attributes from a small set of documents, we observe that \systemname{}'s ranking is noisier than when every document was processed in \systemnamea{}, i.e. an important attribute may be selected by the LLM a few times amongst the $k$ documents. Thus, we show the LLM a union of extracted attributes and prompt it to identify the most useful attributes (prompt in \cite{arora2023evaporate}). 
The frequency-based rank is upweighted if the attribute is in the LLM output.

 \begin{figure}
    \centering
\includegraphics[width=0.48\columnwidth]{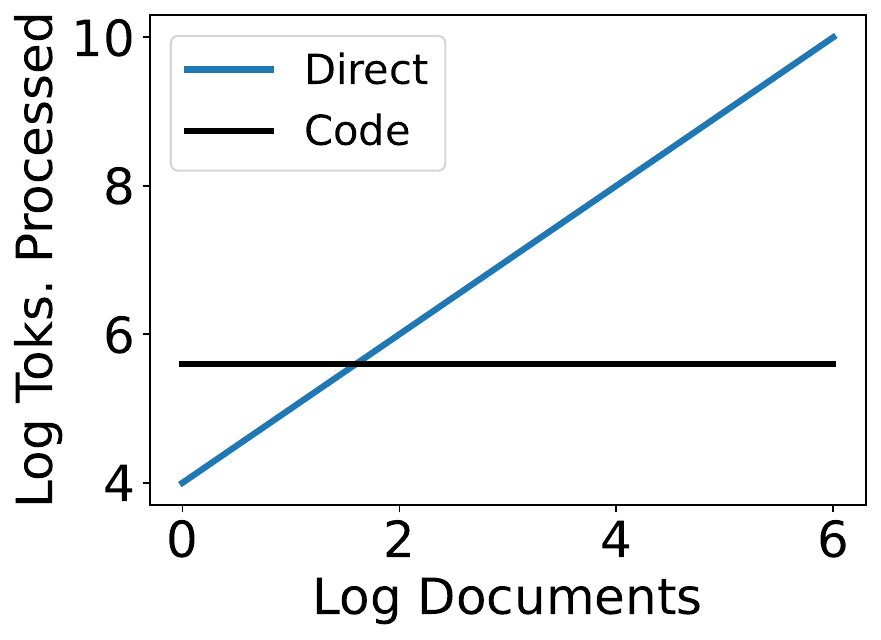}
\includegraphics[width=0.48\columnwidth]{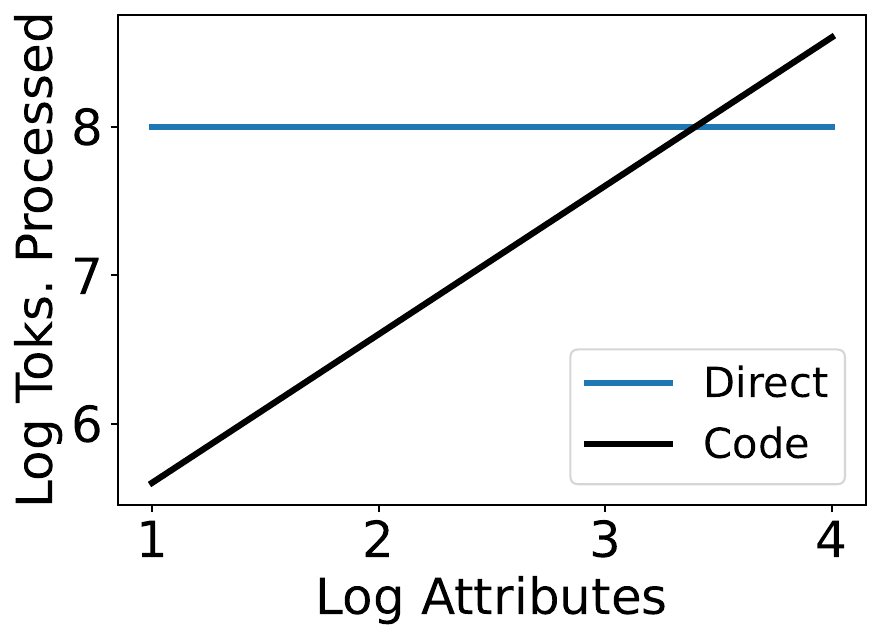}
    \caption[width=\columnwidth]{Tradeoffs between processing the documents via direct prompting (Direct) versus code synthesis (Code). For small data lakes and large numbers of attributes, Direct is sufficient. As the number of documents grows, Code is orders-of-magnitude more efficient. Left is evaluated at 10 attributes, Right at 10K documents, assuming 10K tokens per document.}
    \label{fig:crossinglines}
\end{figure}

\subsubsection{Function Synthesis} 
\label{sec:approach_extract}
Given the attributes $A = \{a_1, a_2 ... a_m\}$, the objective of \systemnameb's second phase is to extract the values of the attributes for each document $d_i \in D$.   
Our key insight, as discussed, is that attribute-values are expressed in similar ways from document to document. To exploit this, instead of processing every document with the LLM to extract values for attribute $a_i$, we propose to use the LLM to \textit{generate code} that can then be reused to process many documents. 

\begin{figure}
    \centering
    \includegraphics[width=0.83\columnwidth]{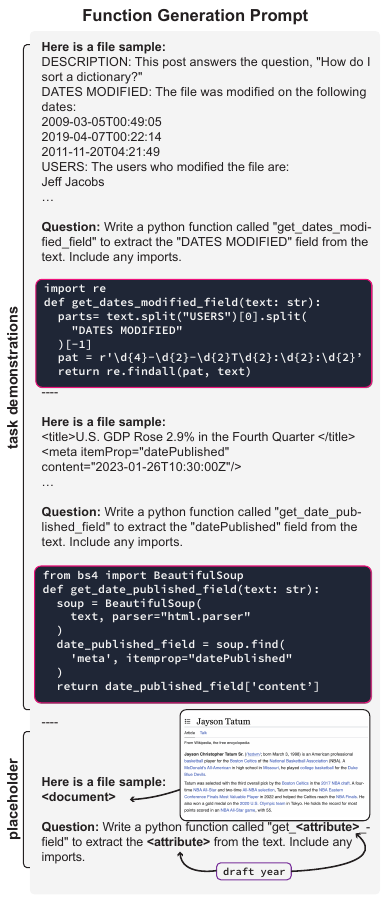}
    \caption[width=\columnwidth]{A representative prompt for function synthesis, containing two data lake agnostic in-context examples.}
    \label{fig:function_prompts}
% \vspace{-7mm}
\end{figure}

Figure \ref{fig:function_prompts} shows an \systemname{} function synthesis prompt. The in-context examples show pairs of text snippets and functions to extract an attribute of interest. \systemname{} searches the data lake via a simple keyword search for document portions that mention $a_i$, and includes this in the prompt. \systemname{} synthesizes functions for attributes following the rank-order of attributes derived during schema synthesis. This means that values of the most relevant (and frequent) attributes as determined by \systemname{} are extracted first. The user can stop the synthesis when desired.

\textbf{Analysis.} We briefly analyze the \systemnameb implementation along the axes of our three desiderata. Results processing the documents with \systemnamea{} are reported in \cref{tab:evaporate_vs_fm_openie} and are discussed in detail in \cref{sec:analysis}.

\textit{Cost.} \cref{fig:crossinglines} demonstrates the asymptotic differences in cost between \systemnamea{} and \systemnameb{}. 
\systemnameb is asymptotically more efficient as a function of the number of documents: the number of LLM calls required with function generation is proportional to the number of attributes, not the number of documents. The crossover point is at $\sim$40 documents. Meanwhile, \systemnamea{} has the potential to extract multiple attributes from the in-context document per inference call, while \systemnameb{} requires generating new functions for each attribute. Thus, the cost of \systemnameb grows with the number of attributes, while the cost of \systemnamea{} approach is constant. The crossover point is at $\sim$2,500 attributes (\cref{fig:crossinglines}).

\textit{Quality.} The tables generated by \systemnameb are on average 21.9 pair F1 points worse than those
produced using \systemnamea on the SWDE datasets (Table \ref{tab:evaporate_baselines}). This suggests that there is a cost-quality tradeoff between the two implementations, since \systemnameb is much cheaper.

\subsection{\systemnamec} 
\label{sec:approach-codeplus}
In this section we discuss an extension of \systemnameb, which enables significant quality improvements while keeping costs low. 
This implementation, which we call \systemnamec, synthesizes \textit{many} candidate functions and ensembles their extractions using weak supervision. 
We decompose the task into three parts:

\begin{enumerate}
    \item \textbf{Schema identification.} (\cref{sec:approach_schema}) Same as in \systemnameb.
    \item \textbf{Function synthesis.} (\cref{sec:approach_diverse}) Same as in \systemnameb, except instead of generating a single function per attribute, we generate many \textit{candidate functions}. Below we describe techniques to encourage diversity among candidates. 
    \item \textbf{Function Aggregation.} (\cref{sec:approach_aggregation}) The synthesized candidate functions have varying qualities and coverages, making them unreliable. We then introduce a weak supervision (WS) based algorithm to aggregate over their different predictions for the attribute values across documents. 
\end{enumerate}

\subsubsection{Synthesizing Diverse Candidate Functions} 
\label{sec:approach_diverse}
We find that the quality of LLM-generated functions varies significantly depending on the document chunk and in-context examples used in the prompts. To address the variability in function quality, we adopt the strategy we previously proposed in \citet{arora2022ama}. This strategy curates multiple diverse prompt templates for the same task (i.e. multiple function generation prompts in the style of \cref{fig:function_prompts}) and prompts the LLM with each in turn to produce a diverse set of \textit{function candidates} $F = \{f_1, f_2, ... f_{k}\}$. 

\systemname{} permits the use of multiple function generation prompts. We use $P_A$ and $P_B$ (included in \cite{arora2023evaporate}) in this work. $P_A$ has zero in-context examples and a task description that encourages the LLM to use regex. $P_B$ has two in-context examples and a task description that encourages the LLM to import and use any Python library. We find that neither consistently outperforms the other. $P_A$ produces higher quality functions on $69\%$, $45\%$, $60\%$, $91\%$, and $31\%$ of attributes on the 8 SWDE Movie, 5 SWDE University, FDA reports, Enron, and Wikipedia player pages settings respectively. Designing a single ``perfect'' prompt can be challenging so \systemname{} aggregates results from multiple prompts.

\subsubsection{Aggregating Candidate Functions}
\label{sec:approach_aggregation}

Next, we discuss how to combine the aggregations of the candidate functions. 

\textit{Background: Methods for Unsupervised Aggregation}
Because we lack ground truth labels in our setting, it is not possible to directly evaluate the quality of the candidate functions. A popular unsupervised aggregation strategy is to take the Majority Vote (MV) across function outputs \cite{wang2022selfconsistency}. Formally, MV treats the functions as independent of one another and assigns equal weight to all function outputs. However, the functions are not of equal quality --- over 40\% of synthesized functions result in less than 25 Text F1 in extraction quality. Therefore, \systemname{} uses weak supervision (WS), a popular standard statistical framework for modeling the accuracies and correlations between noisy sources of information without any labeled data \cite{ratner2017snorkel, fu2020fast}. In WS, we learn a \textit{label model} that is parametrized by the accuracies and correlations of the candidate functions. WS is widely used in industry \cite{ratner2017snorkel}. 

Unfortunately, existing WS setups make the following assumptions that do not apply in our setting. The standard setup assumes \textit{human-designed} functions while our setting uses \textit{machine-generated} functions that output non-standardized extracted text.
% The following standard assumptions do not hold in our setting:

\begin{enumerate}
    \item \textbf{Assumption 1: Functions will abstain on examples where they do not apply \cite{ratner2017snorkel, fu2020fast}.} The attribute value returned by a function for a document could be \verb|null| for two reasons:  (1) the attribute does not exist in the document (e.g. a Wikipedia page may be missing a college attribute since the players did not attended college) or (2) the attribute exists but the function was not sophisticated enough to extract it (e.g. the  \verb|product code| attribute could start with a lowercase ``k'' of uppercase ``K'' in FDA reports, but the particular function is only designed to extract for lowercase, resulting in empty strings for uppercase documents). Note that in (1), if the function outputs a value, it has low precision and we would want to ignore the function. Note that in (2), the function has high precision and ideally our system learns to utilize such functions \textit{selectively}. Unfortunately, it is challenging to determine whether the function outputs \verb|null| for reason (1) or (2) in our setting, whereas in the traditional WS setup with human-provided functions, humans specify this logic directly (e.g. ``If the email has a URL, ``vote'' that it contains spam, otherwise abstain'' \cite{smith2022lmws}). 
    \item \textbf{Assumption 2: Functions are correlated with the gold label $y$ at better than random performance \cite{ratner2017snorkel, fu2020fast}}. While this is reasonable when functions are human-designed, \systemname{} uses machine-generated functions. We find 51\% of generated functions are below 50 Text F1. 
    \item \textbf{Assumption 3: Weak supervision is typically applied to tasks with well defined classes in a classification setting \cite{ratner2017snorkel, fu2020fast}.}  In our case, the output of the functions are extracted text, and thus there is a virtually unconstrained output space of possible extractions that vary from document to document (e.g. NBA players have varied \verb|date of birth| values). The \textit{number} of unique extractions collected by the functions can also differ across documents.
\end{enumerate}

% \vspace{-2mm}
We propose the following approach to be able to leverage WS. Let $D_{eval}$ be a small sample of documents from the data lake $\mathcal{D}$. We have the set of generated functions $F$ and LLM $\mathcal{F}$.

\textbf{Handling function abstentions.} To estimate the probability that an empty output from a function is an abstension, we propose to measure the fraction $e$ of the $D_{eval}$ documents for which $\mathcal{F}$ extracts a value. Intuitively, when $e$ is high, our prior should be that the attribute appears in a large fraction of documents, so we should assume functions are \textit{abstaining} when they output empty values. When $e$ is low, the attribute appears in few documents, so we should assume the functions are \textit{predicting} empty values. We can use $e$ to guide both our function evaluation and downstream aggregation. Note that it is possible for $\mathcal{F}$ to abstain or hallucinate values, affecting the estimate of $e$.

\begin{algorithm}[t]
\caption{Function Aggregation (from \systemnamec)}
\begin{algorithmic}[1]
\STATE \textbf{Input:} Documents $\mathcal{D}$, candidate functions $F$, LLM $\mathcal{F}$. \newline \textbf{Output:} Predicted extractions ${\hat{y}_i, ..., \hat{y}_n}$ for documents.
\STATE \textbf{Collect sample predictions} Sample $\mathcal{D}_{eval} \subset \mathcal{D}$ and apply the functions $f_j \in F$ and LLM $\mathcal{F}$ to obtain $\hat{y}_{ij}$ and $\hat{y}_{i\mathcal{F}}$ for document $d_i$.
\STATE \textbf{Handle abstensions}: For empty $\hat{y}_{ij}$, we need to determine if they represent \textit{function abstensions} or \textit{predictions} that $d_i$ has no-value for the attribute. Use $\mathcal{F}$ to decide between cases: compute $e$ as the fraction of $d_i \in \mathcal{D}_{eval}$ with non-empty $\hat{y}_{i\mathcal{F}}$
% fraction $e$ across 
\STATE \textbf{Score functions}: Compute a score $\hat{a}_j$ for $f_{j}$ using metric function $m(\cdot)$ based on $e$. \\
    \textbf{if $e > \tau$ then} \\  \setlength\parindent{24pt} $\hat{a}_j = \sum_{i=1}^{i=n} m(\hat{y}_{i\mathcal{F}}, \hat{y}_{ij}) | \hat{y}_{i\mathcal{F}} \neq \emptyset$ \\
    \noindent \textbf{else} \\
    \setlength\parindent{24pt} $\hat{a}_j = \sum_{i=1}^{i=n} m(\hat{y}_{i\mathcal{F}}, \hat{y}_{ij})$ \\
\STATE \textbf{Filter low quality functions} Remove $f_j \in F$ with $\hat{a}_j \leq 0.5$ to create $F'$.
\STATE \textbf{Collect votes} 
Apply $f \in F'$ to all $d_i \in \mathcal{D}$ to collect ``votes'' for the attribute-value in $d_i$. Post process empty votes as \textit{abstensions} or no attribute \textit{predictions} depending on $e$. 
\STATE \textbf{Aggregation} Use weak supervision to obtain the final prediction $\hat{y}_i$ given the function votes $\{\hat{y}_{ij}|f_j \in F'\}$.
\end{algorithmic}
\label{alg:approach}
\end{algorithm}

\textbf{Handling functions with worse than random quality.} 
We propose to utilize the extractions fro $\mathcal{F}$ on a small set of documents $D_{eval}$ (e.g. we use $|D_{eval}| \leq 10$)
as an estimate of the ground truth extractions for those documents.
We can then estimate the quality, $\hat{a}_j$, of function $f_j$ by comparing its outputs against the outputs of $\mathcal{F}$ on document $d_i \in \mathcal{D}_{eval}$. If we are in the low $e$ regime, we should evaluate the outputs on all $d \in \mathcal{D}_{eval}$. In the high $e$ regime, we should evaluate the outputs on only the $d \in \mathcal{D}_{eval}$ for which $f_j$ extracted a value. We finally filter $f_j$ if $\hat{a}_j \leq 0.5$, where $0.5$ derives from the typical WS assumptions \cite{ratner2017snorkel, varma2018snuba, fu2020fast}.

Note that $\mathcal{F}$ is an LLM with its own error rate $e$, affecting the estimate $\hat{a}_j$. We theoretically study the impact of $e$ on the label model learned via WS. We provide the proof in \cite{arora2023evaporate}.

\textit{Proposition 1: We have $m$ functions with empirical accuracies $\hat{a}$, evaluated against noisy labels with error rate $e$, the function accuracies estimated by the weak supervision label model are $\Tilde{a}$, and the measured error is below some threshold $\epsilon$. Then, if each function labels a minimum of $$n \geq \frac{1}{2(\gamma-\epsilon - e)}\log(\frac{2m}{\delta})$$ datapoints, the weak supervision label model will succeed in learning accuracies such that $||a^* - \Tilde{a}||_{\infty} < \gamma$ with a probability $1 - \delta$.}

\textbf{Handling unconstrained output spaces.} The $k$ generated functions can produce $[0..k]$ unique prediction votes for a single unlabeled document $d_i$, and the number of unique votes can differ from document $d_i$ to $d_j$. Therefore, for each $d_i \in \mathcal{D}$, we  bucket the unique votes and take the $b$ buckets representing the most frequently occurring votes. The votes for functions that outputted values outside the top-$b$ are marked as abstensions. If the number of unique votes is $< b$, placeholder values are inserted into the top-$b$. Finally, as the ``classes'' differ across documents, we introduce a constraint to the objective function encouraging the class-conditional accuracies to be equal.  

After addressing these assumptions, we can leverage prior approaches to aggregate the noisy extractions from the function candidates into higher-quality extractions as in \cite{ratner2017snorkel, arora2022ama}. Under WS, the output of each function is viewed as a ``vote'' for the true label and the objective is to construct a latent graphical model to account for the varied accuracies and correlations amongst the functions, without access to any labeled data. Our aggregation method is summarized in \cref{alg:approach}.

\textbf{Analysis.} We briefly analyze the \systemnamec implementation along the axes of our three desiderata. Results processing the documents with \systemnamec{} are reported in \cref{tab:evaporate_vs_fm_openie} and are discussed in detail in \cref{sec:analysis}.

\textit{Cost.} As with \systemnameb, the number of tokens processed by the LLM in \systemnamec is fixed with respect to the number of documents. 
\cref{fig:crossinglines} demonstrates the asymptotic differences in cost between \systemnamea and \systemnameb. The number of tokens that must be processed by the LLM grows only by a constant factor: the number of function candidates generated. The user can set this number to balance cost and quality.

\textit{Quality.} Of the three implementations, \systemnamec produces the highest quality tables. \systemnamec{} outperforms \systemnamea by 12.1 F1 points (22\%) on average, while using far fewer computational resources. Using function aggregation leads to an improvement of 25.1 F1 points over \systemnameb.

%% file: sections/analysis.tex
\label{sec:analysis}
We now evaluate \systemname{}, validating the following claims:

\begin{itemize}
    \item \textbf{Function synthesis enables asymptotic cost reductions for processing data with LLMs.} There has been significant recent interest in developing various data management applications with LLMs \cite{khattab2022demonstrate, chen2023symphony, narayan2022can, Lai2022DS1000}. Prior work directly processes data with the LLM. \systemnamec{} reduces the number of tokens the LLM needs to process by 110x relative to \systemnamea{}.
    \item \textbf{Function synthesis + aggregation results in higher quality than direct extraction.} Despite the fact that \systemnamea{} processes each document with the LLM directly, \systemnamec{} performs 10.1 F1 points (18\%) better on average. Based on comparisons with \systemnameb, which only synthesizes one function, we show that function aggregation is key in enabling the improvements.
    \item \textbf{\systemname achieves higher quality than state-of-the-art baselines, while exposing a more general interface.} \systemnamec expresses tasks via merely \textit{six} natural language prompts (all provided in \cite{arora2023evaporate}) and uses no training. Yet, it exceeds SoTA systems by 3.2 F1 (6\%) points when generating tables from scratch and 6.7 points (10\%) when extracting pre-defined gold attributes. Meanwhile, it supports a broader range of settings than any of these baselines.
   \item \textbf{The identified tradeoffs hold across language models.} We evaluate on four models from three unique providers \cite{openai, askell2021anthropic, lieber2021jurassic}. We find  \systemnamea{} and \systemnamec{} remain competitive in quality across LLMs.
\end{itemize}

\subsection{Experimental Setup} 
\label{sec:baselines}
We primarily evaluate \systemname{} on the end-to-end task of \textit{structured view generation}. For the purpose of comparison to prior work, we also evaluate on the sub-task of \textit{closed information extraction}. We first define these tasks, their metrics, and the baselines. We then provide implementation details for \systemname{}.

\textbf{Structured view generation task.} This captures the end-to-end task of identifying the schema and populating the output table. This task is often discussed as a vision system \cite{cafarella2007navigating}, and given the difficulty of this task, there are limited comparable works. We therefore compare to the closest line of work, OpenIE systems, where the task is to extract all facts from documents \cite{banko2007openie, niklaus2018openie}. We compare to two sets of baselines: (1) \citet{lockard2019ceres, lockard2020zeroshotceres, deng2022domlm} for \textsc{HTML}-specific OpenIE, and (2) \citet{kolluru2020openie} for generic unstructured text. The former models explicitly use the \textsc{HTML}-DOM tree structure to process the page, assuming attribute values are leaf nodes, and explicitly train on documents from the domain of interest. The latter class of systems first label sentences using linguistic tools (\textit{i.e.} dependency parsers, part of speech taggers, and named entity taggers), and fine tune LLMs over these features to perform the task \cite{zhou2022survey}. 

\textit{Metrics.} The standard metric is Pair F1 \cite{lockard2019ceres, deng2022domlm}, an F1 score applied to the predicted vs. gold sets of tuples of the form (document ID $d_i$, attribute $a_j$, value $r_i,j$). The tuple must exactly match a tuple in the ground truth to be marked correct. Since \systemname{} ranks the attributes and generates functions in this order, for fair comparison, we report OpenIE scores for all tuples up to $k$ attributes, where $k$ is the number of gold attributes for the setting. 
We note that the prior systems extract all-or-no tuples, in contrast. 

\textbf{Closed information extraction task.} This captures the setting where the user provides a pre-defined schema and \systemname{} is used to populate the table. We compare to state-of-the-art approaches for closed IE including: (1) \citet{lockard2019ceres, lockard2020zeroshotceres, deng2022domlm} for \textsc{HTML}-specific ClosedIE and (2) \citet{he2021deberta, clark2020electra} for generic unstructured text. The former models explicitly use the \textsc{HTML}-DOM tree structure to process the page, assuming attribute values are leaf nodes, and explicitly train on documents from the test domain. The latter are pretrained LLMs that have been fine tuned on massive amounts of labeled \verb|(attribute, value)| pairs \cite{rajpurkar2022squad}. We report ClosedIE results using the Text F1 metric on a value-by-value basis across each document.

\textbf{\systemname{} Implementation Details.} In the following experiments, we instantiate \systemname{} with currently popular, LLM APIs. Experiments in \cref{sec:compare-davinci,sec:eval} use \verb|text-davinci-003| from OpenAI. In \cref{sec:compare-models}, we evaluate additional LLMs from three model providers. For experiments, we use 10 sample documents per data lake for the schema synthesis, function synthesis, and function verification. 
We apply \cref{alg:approach} over the top-10 scoring functions that are synthesized for each attribute and data lake. The prompts remain constant across data lakes and models. In  \cite{arora2023evaporate}, we provide ablations that show how the system's quality changes as we vary the number of sample documents and top-$k$ functions. 

When the measuring cost for alternate implementations of \systemname{}, we compute total number of tokens processed by the LLM to perform the end-to-end task (\textit{i.e.} the sum of the number of tokens in the prompt and model generation). We use this metric because the wall-clock time and dollar cost of a model fluctuate, but both should be proportional to the number of tokens processed.

% \vspace{-2mm}
\subsection{Evaluation Settings} 
\label{sec:evaluation-settings}
We evaluate \systemname{} on 16 settings representing a range of real-world data lakes. First, we use a benchmark suite of 13 Movie and University websites to compare \systemname to state-of-the-art information extraction systems \cite{hao2011swde, lockard2019ceres, deng2022domlm}. Next, to evaluate on more unstructured data (\textit{i.e.} non-\textsc{HTML}) we turn to: \textbf{Enron} a corporate email corpus that has been analyzed in over three thousand academic papers \cite{klimt2004enron, heller2017enronnyt, arora2022reasoning}, \textbf{FDA 510(k)} reviews for premarket notification submissions for medical devices, which have been the subject of multiple important research studies \cite{wu2021medai,zuckerman2011devicerecalls}, and  \textbf{NBA} Wikipedia pages for NBA players, which include more complex \textsc{HTML} than the existing benchmarks \cite{deng2022domlm}. We release the benchmarks and provide additional details in \cite{arora2023evaporate}. Here we briefly describe the properties we aim to study with each setting:
\begin{enumerate}
    \item \textbf{Benchmark Suite: SWDE Movies \& Universities} SWDE is the standard benchmark for document-level IE in prior work \cite[inter alia.]{hao2011swde, lockard2019ceres, lockard2020zeroshotceres, deng2022domlm}. There are 8 sets of webpages for Movies (e.g. IMDB) and 5 sets of webpages for Universities (e.g. US News). For each website, the benchmark contains 1063-2000 pages and annotations for 8-274 attributes. We use SWDE to compare to the state-of-the-art and test on a range of attribute types, e.g. simpler Movie ``runtime'' through complex Movie ``cast'' and popular Movie ``director'' through infrequent ``second assistant director''. 
    \item \textbf{Complex \textsc{HTML}: NBA} As SWDE attributes always occur in separate leaf nodes of the \textsc{HTML}-DOM tree, we use NBA Player Wikipedia pages to evaluate on more complex \textsc{HTML}. E.g., the \verb|NBA| draft attribute contains the draft round, year, pick number, and team by which the player was selected. We evaluate on 100 randomly selected player pages (spanning the 1940s-present) and 19 attribute annotations.
    \item \textbf{Unstructured Text: Enron and FDA} We observe a lack of existing benchmarks for document-level IE over unstructured text --- intuitively, this setting has been challenging with prior generations of models due to the lack of \textit{any} grounding structure whatsoever (\textit{i.e.} recall current systems rely on \textsc{HTML}-DOM elements or sentence-level NER, dependency, and POS tags). 
    We turn to the Enron and FDA settings described above. The Enron setting contains 15 gold attributes and 500k documents. The FDA setting contains 16 gold attributes and 100 PDF documents, which are up to 20 pages long, randomly sampled from FDA 510(k).
\end{enumerate}

\vspace{-2mm}
\textbf{Dataset Protocols} Because the baselines we compare against on SWDE require training data, they perform website-wise cross validation (\textit{i.e.} train on some websites and evaluate on others). They do this for several combinations such that every website appears in the evaluation set. \systemname, in contrast, does not require any training data. We simply evaluate \systemname{} on all websites so that our method is evaluated on the same examples as the baselines.

\begin{figure}
    \centering
    \includegraphics[width=0.98\linewidth]{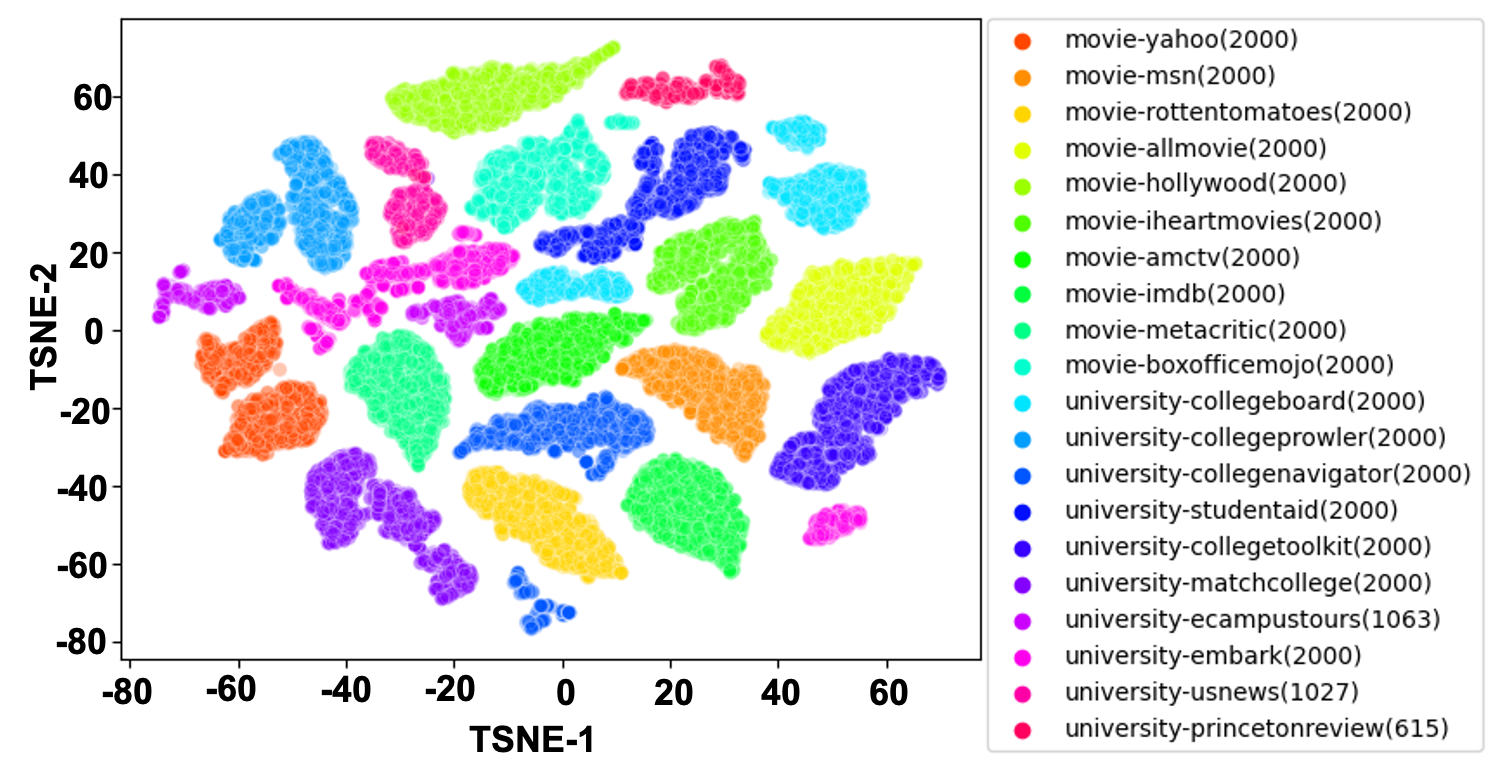}
    \vspace{-2mm}
    \caption[width=\linewidth]{T-SNE Visualization of documents in the SWDE dataset. T-SNE is performed on the first 16 principal components of TF-IDF vectors. Colors indicate the source website.}
    \label{fig:clusters}
    \vspace{-2mm}
\end{figure}

We process each dataset with \systemname separately, to match the protocol used by the baselines~\cite{deng2022domlm}. However, we may not have access to the sources of the documents in a real-world data lake -- they may be mixed together. Using a standard TF-IDF vectorizer and K-means clustering to the mixture of documents, we verify we can perfectly recover the document sources without any labeled data or supervision (Figure \ref{fig:clusters}, details in \cite{arora2023evaporate}). Intuitively, clustering semi-structured data may be simple due to the rich formatting.

% \vspace{-2mm}
\subsection{Comparing \systemname{} to Baselines}
\label{sec:eval}

First we validate that \systemname{} outperforms baselines, defined in \cref{sec:baselines}, both in generality (i.e. the flexibility to support data from different domains and formats) and quality. We then compare the efficiency of \systemnamec vs. the baselines.

\subsubsection{Quality and generality comparisons} 

\paragraph{Systems for semi-structured text} Shown in \cref{tab:evaporate_baselines}, \systemname{} outperforms the state-of-the-art on SWDE. We compare to the metrics reported the baseline works. 
 Recall \systemname{} uses no training whatsoever and can be applied across document formats (\textsc{HTML}, \textsc{PDF}, \textsc{TXT}). In contrast, the baselines are limited to \textsc{HTML} and explicitly perform supervised learning using labels from webpages within the Movie and University domains respectively \cite{lockard2020zeroshotceres, deng2022domlm}. E.g., \citet{deng2022domlm} assumes attribute values are the leaf-nodes of the \textsc{HTML}-DOM tree and thus does not work on non-\textsc{HTML}.

The baseline systems restrict scope to attributes that are specifically mentioned in the \textsc{HTML} \verb|<body>| text, even though attributes are frequently mentioned in the \textsc{HTML} header (e.g. within \verb|<title>| elements) and tags (e.g. \verb|<a href='year/2012'>|). We validate \systemname{} can identify and extract attributes mentioned anywhere in the document. We extend the SWDE benchmark to include the attributes scattered throughout the full \textsc{HTML} and find \systemname{} achieves 52.2 and 49.0 on Movies and University respectively on the more challenging setting. We release the new annotations. 

\input{tables/main_quality}
\input{tables/top_level_results}

\input{tables/evaporate_vs_fm_openie}

\paragraph{Systems for unstructured text} We are not aware of strong baselines that apply beyond \textsc{HTML} document formats. The most relevant baseline is the OpenIE6 system for performing OpenIE over any unstructured text from \citet{kolluru2020openie}. We find the system only handles well formatted sentences and struggles to extend to heterogeneous data types. We find that even when documents contain full sentences, the system extracts an extremely large set of relations and does not enforce consistent extractions across documents. For instance, on a sample FDA 510(k) document, OpenIE6 extracts 427 relations with 184 relations having a confidence level at 0.99. We include a detailed error analysis in \cite{arora2023evaporate}.

\subsubsection{Efficiency comparisons} We compare the efficiency of \systemname{} with the (estimated \cite{sizesOpenAI}) 175B parameter OpenAI model vs. the baseline, which uses a pretrained 125M parameter RoBERTa model \cite{deng2022domlm}, decomposed in terms of pretraining, fine-tuning, inference, and parameter  (memory) cost. Using FLOPS as reported in the OpenAI reference work \citet{brown2020language}, for data settings with $n$ documents and $m$ attributes, the costs are: 
\begin{itemize}
    \item \textbf{RoBERTa} The model is 125M parameters, total pretraining FLOPS is 1.50E+21, and inference FLOPS per token is $2 \times \text{Number of Parameters}$, which is $0.250$ GFLOPS. The total inference cost is computed by $$n \times \frac{\text{tokens}}{document} \times 0.250 \text{ GFLOP}$$
    \item \textbf{GPT3-175B} The model is 175B parameters, total pretraining FLOPS is 3.14E+23, and inference FLOPS per token is $2 \times \text{Number of Parameters}$, which is $350$ GFLOPS. The total inference cost is computed by $$m \times P \times \frac{\text{tokens}}{chunk} \times 350 \text{ GFLOP}$$ 
    where $P$ is the number of prompts per attribute. Note that for our evaluated implementation $P \approx 10c$ since function generation is performed on $10$ documents.
\end{itemize}
\systemname{} uses a model with 1,400x more parameters and 300x higher pretraining cost. However, users of the baselines likely need to locally fine-tune and host the models. The inference costs of the baseline and \systemname{} on our datasets are in the same order of magnitude (summarized in \cite{arora2023evaporate}). The extended cost comparison and analysis are provided in \cite{arora2023evaporate}.  Users should select a method depending on the data setting, i.e. the number of documents and attributes. We note that \systemname allows users to tradeoff quality and efficiency by changing the underlying language model to smaller variants. 

% \vspace{-2mm}
\subsection{Comparing Implementations of \systemname{}} 
\label{sec:eval-implementations}

This work proposes a fundamental tradeoff space between directly processing data workloads with LLMs vs synthesizing code that does the processing. We first discuss the tradeoffs for a fixed LLM (\verb|text-davinci-003|), which is the current best-in-class LLM \cite{liang2022helm} (\cref{sec:compare-davinci}), and next across a range of LLMs trained by three distinct model providers (\cref{sec:compare-models}). 

% \vspace{-1mm}
\subsubsection{Tradeoffs between \systemname{} Implementations}
\label{sec:compare-davinci}
As detailed in \cref{sec:approach}, the base routine (``\systemnamea{}'') in \systemname{} entails directly processing documents with the LLM, while the optimized routine (``\systemnameb{}'') synthesizes functions for processing. Next we evaluate these along our desiderata.

\setlength\parindent{12pt} \textbf{Generality is maintained.} LLMs take text as input and provide text as output --- this unified natural language interface means \systemnamea{} and \systemnameb{} can ingest any document format without additional engineering. \textit{Critically, our results with \systemname{} require no user effort, no training whatsoever, and no customization when applied to the 16 different settings}. \\
% \vspace{-1mm}

\setlength\parindent{12pt} \textbf{Asymptotic cost reduction.} \cref{fig:crossinglines} demonstrates the asymptotic differences in cost between directly processing the data lake with \systemnamea{} vs. with \systemnamec{}. (\cref{fig:crossinglines} Left) \systemnamea{} is asymptotically more efficient as a function of the number of documents in the data lake. The number of LLM calls required with function generation is proportional to the number of attributes to be extracted, not the number of documents. The crossover point is at $\sim$40 documents. 

(\cref{fig:crossinglines} Right) \systemnamea{} can extract multiple (\textit{i.e.} every) attribute in the in-context documents in a single inference call, while \systemnamec{} synthesizes new functions for each attribute. Thus, the cost of function synthesis grows with the number of attributes, while the cost of \systemnamea{} is constant. The crossover point is at $\sim$2,500 attributes. 

Empirically across our settings, \systemnamec{} realizes a 110x average reduction in the number of tokens the LLM needs to process (assuming 10k documents per setting and 378$\times$ given the true benchmark sizes)  in the number of tokens the LLM must process compared to \systemnamea{} (\cref{tab:evaporate_vs_fm_openie}). Further, data lakes are constantly changing and functions can be reused while \systemnamea{} would need to be re-run, multiplying the cost. \\

% \vspace{-3mm}
In runtime, we observe that the generated functions are efficient in processing the documents. For example, over the 9,500 function runs (from 95 functions evaluated on 100 documents each) in the FDA 510(k) setting, we find that the average time to run one function over one document is 0.00025s on a 2 CPU machine. \\

% \vspace{-2mm}
\setlength\parindent{12pt} \textbf{Improved quality and reliability.} Even though \systemnamea{} directly processes each document with the LLM, \systemnamec{} surprisingly performs 10.1 F1 (18\%) better (\cref{tab:evaporate_vs_fm_openie}).

\textit{What are the failure modes of \systemnamea{}?} The method yields inconsistent generations. On the Medical FDA report setting:(1) The LLM misses an average of 4.4  attributes that are present in the gold schema (27.5\% of gold attributes) per document. Among the gold attributes that are missed, all \textit{are extracted} in at least one document. (2)  Further, the LLM outputs an average of 9.7 attributes or values that are not explicitly mentioned in the documents. (3) Finally, attributes are reworded in diverse ways across documents --- the attribute \verb|classification| is extracted in 4 different ways across the sample of 10 documents (i.e. ``classification'', ``device classification'', ``regulatory information'', missing). Since the error modes are quite varied, it is unclear how to improve quality.

\paragraph{Why does \systemnamec{} improve quality?}
We validate that our Algorithm \ref{alg:approach} for selecting and aggregating functions leads to the quality improvements over \systemnamea{}. 

\textbf{Synthesizing diverse functions} We find that using diverse prompts helps address the lack of reliability in function synthesis. To synthesize functions, \systemnamec{} uses a prompt template that includes one-to-two in-context examples and a placeholder for the inference example, i.e. document text (\cref{fig:function_prompts}). We can produce multiple prompts in the template by swapping the in-context examples or sampling more documents (change the inference example). We find both means of increasing diversity benefit quality:
\begin{itemize}
    \item \textbf{In-context demonstrations} Our implementation (\cref{tab:evaporate_openie}) instantiates two prompts by swapping in-context demonstrations , $P_A$ and $P_B$. Quality using $P_A$ or $P_B$ alone is 8.5 and 8.0 F1 points worse than using both to synthesize functions on SWDE Movie and SWDE University respectively.
    % respectively highlighting the benefit of prompting with diverse examples. 
    \item \textbf{Inference documents} Using three versus five sample documents in the prompts for \systemnamec{}, the ClosedIE and OpenIE quality improve by 6.8 F1 points (9\%) and 6.5 F1 points (14\%) respectively, averaged across the 16 settings. 
\end{itemize}

\textbf{Estimating function quality using the LLM.} In \cref{tab:ws_agg_ablation}, we first evaluate the two unsupervised aggregation baselines in prior work off-the-shelf: Majority Vote (MV) and Weak Supervision (WS) \cite{ ratner2017snorkel, wang2022selfconsistency, arora2022ama}. Next we measure the effect of filtering functions and handling abstentions as proposed in \cref{alg:approach}. 

In \cref{tab:ws_agg_ablation}, we observe WS with filtering provides a consistent boost across settings compared to WS --- 7.1 F1 point higher average quality and up to 13.8 F1 points on the SWDE University setting. Additionally handling abstensions leads to a 1.9 F1 point increase in average quality over WS with filtering, with up to 7.8 F1 points on the FDA setting. 
Qualitatively, accounting for abstensions is helpful when attributes are expressed in diverse ways across documents, which is not applicable to all settings such as Enron. 
These results highlight the importance of \systemnamec's aggregation approach for the system's overall reliability. Without \cref{alg:approach}, quality does not improve over \systemnamea{}.

\input{tables/ws_agg_ablation}

%% file: tables/main_quality.tex
\begin{table}[t]
    \caption{Quality of \systemnamec{} evaluated on ClosedIE in Text F1 and OpenIE in Pair F1 using text-davinci-003.}
    \vspace{-3mm}
    \centering
    \resizebox{0.48\textwidth}{!}
    {\renewcommand{\arraystretch}{0.98}
    \begin{tabular}{lc|ccc}
     \toprule
       \multirow{2}{*}{Source (Format)}  &      \multicolumn{1}{p{1.4cm}}{\centering \textsc{ClosedIE}} &
       \multicolumn{3}{p{2.8cm}}{\centering \textsc{OpenIE}} \\
       &   \emph{F1}  &        
      \emph{R}  &   \emph{P} & \emph{F1}  \\
    \midrule
     FDA (\textsc{TXT})   & 80.1 & 62.0 & 68.1 & 64.9 \\
     Enron Emails (\textsc{TXT})   & 93.3 & 80.3 & 94.6 & 86.9 \\
     Wiki NBA (\textsc{HTML})         & 84.7 & 55.7 & 88.2 & 68.2 \\
     SWDE Movie (\textsc{HTML})       & 79.5 & 48.5 & 71.0 & 56.8 \\
     SWDE University (\textsc{HTML})  & 73.7 & 50.9 & 71.4 & 59.0 \\ 
     \hline
     \textbf{Average}  & \textbf{82.3} & \textbf{58.9} & \textbf{78.5} & \textbf{66.7} \\ 
     \bottomrule
    \end{tabular}}
\label{tab:evaporate_openie}
\vspace{-3mm}
\end{table}

%% file: tables/top_level_results.tex
\begin{table}

    \caption{Comparisons to state-of-the-art on ClosedIE in Text-F1 and OpenIE in Pair F1. The baselines train on in-domain documents, while \systemname{} uses no training \cite{deng2022domlm}.}
    \vspace{-3mm}
 % \begin{center}
    % \small
    \begin{tabular}{lcc|cc}
    \toprule
       \multirow{2}{*}{System}  &      \multicolumn{2}{p{2.2cm}}{\centering \textsc{SWDE Movie}} &
       \multicolumn{2}{p{2.5cm}}{\centering SWDE University} \\
    & \emph{Closed}  &   \emph{Open}  &        
      \emph{Closed}  &   \emph{Open}  \\
    \midrule
    
    ZeroShot Ceres \cite{lockard2020zeroshotceres}  & -& 50.0 & - & 50.0 \\
    RoBERTa-Base & 49.3 & 35.6 & 36.6 & 38.0 \\
    RoBERTa-Structural & 47.7 & 39.9 & 46.5 & 42.3 \\
    DOM-LM \cite{deng2022domlm} & 71.9 & 54.1 & 68.0 & 55.2 \\
    \midrule
    \systemnamea{} & \textbf{84.4} & 45.2 & 72.6 & 53.8 \\
    \systemnameb{} & 55.0 & 33.0 & 40.5 & 22.2 \\
    \systemnamec{} & 79.5 & \textbf{56.8} & \textbf{73.7} & \textbf{59.0} \\
    \bottomrule
    \end{tabular}
    \normalsize
    \label{tab:evaporate_baselines}
% \end{center}
\vspace{-4mm}
\end{table}

%% file: tables/evaporate_vs_fm_openie.tex
% \usepackage{multirow}
\begin{table*}[]
\caption{Quality (OpenIE Pair F1) and cost (number of tokens processed by the LLM) for producing the structured views. We compare the direct prompting and code synthesis implementations using text-davinci-003. \systemnamec{} is evaluated on the full datasets, while \systemnamea{} is evaluated on a randomly sampled proportion of documents due to the cost (20\% of FDA and Wiki, 2\% of SWDE, and 0.0002\% of Enron, given the respective sizes).}
\vspace{-3mm}
\begin{tabular}{lcccccccc}
\hline
\multirow{3}{*}{Source (Format)} & \multicolumn{3}{c}{\systemnamea{}}                               & \multicolumn{3}{c}{\systemnamec{}}                            & \multicolumn{2}{c}{Relative Performance}                   \\
                                 & Quality       & \multicolumn{2}{c|}{Cost / 10K Documents}          & Quality       & \multicolumn{2}{c|}{Cost / 10K Documents}       & \multirow{2}{*}{Quality} & \multirow{2}{*}{Cost Reduction} \\
                                 & F1            & Tokens (M)   & \multicolumn{1}{c|}{Cost (\$)}      & F1            & Tokens (M)   & \multicolumn{1}{c|}{Cost (\$)}   &                          &                                 \\ \hline
FDA (TXT)                        & 45.5          & 145.6        & \multicolumn{1}{c|}{2,900}          & 62.8          & 1.9          & \multicolumn{1}{c|}{38}          & +17.3                    & 77x                             \\
Enron Emails (TXT)               & 93.8          & 21.2         & \multicolumn{1}{c|}{425}            & 86.9        & 0.6          & \multicolumn{1}{c|}{12}          & -6.9                     & 35x                             \\
Wiki NBA (HTML)                  & 44.8         & 650.1        & \multicolumn{1}{c|}{13,000}         & 68.2          & 3.0          & \multicolumn{1}{c|}{60}          & +23.4                    & 217x                            \\
SWDE Movie (HTML)               & 45.2         & 282.9        & \multicolumn{1}{c|}{5,660}          & 56.8          & 2.3          & \multicolumn{1}{c|}{46}          & +11.6                    & 123x                            \\
SWDE University (HTML)           & 53.8       & 190.1        & \multicolumn{1}{c|}{3,800}          & 59.0        & 1.9          & \multicolumn{1}{c|}{38}          & +5.2                     & 100x                            \\
\hline

\textbf{Average}                 & \textbf{56.6} & \textbf{258} & \multicolumn{1}{c|}{\textbf{5,157}} & \textbf{66.7} & \textbf{1.9} & \multicolumn{1}{c|}{\textbf{39}} & \textbf{+10.1}           & \textbf{110x}                   \\ \hline
\end{tabular}
\label{tab:evaporate_vs_fm_openie}
\vspace{-3mm}
\end{table*}

%% file: tables/ws_agg_ablation.tex
\begin{table}[t]
    \caption{Quality under alternate approaches of aggregating the synthesized functions. Baselines are in the left columns: Majority Vote (MV) and Weak Supervision (WS). Components of Algorithm \ref{alg:approach}  are in the right columns: ``Abstain'' accounts for abstensions and ``Filter''  filters low quality functions.}
    \vspace{-3mm}
    \centering
    \resizebox{0.48\textwidth}{!}{\renewcommand{\arraystretch}{0.98}
    \begin{tabular}{lcc|cc}
     \toprule
       \multirow{2}{*}{Source}  &   
       \multirow{2}{*}{\centering \textsc{MV}} &   
       \multirow{2}{*}{\centering \textsc{WS}} &
       \multicolumn{1}{p{1cm}}{\centering \textsc{WS}} & 
       \multicolumn{1}{p{2cm}}{\centering \textsc{WS}} \\
       &  
       &  
       & Filter 
       & Abstain $+$ Filter\\
    \midrule
     FDA (TXT)             & 52.9 & 51.1 &  55.0 & 62.8 \\
     Enron Emails (TXT)    & 81.4 & 82.7 &  86.9 & 86.9 \\
     Wiki NBA (HTML)        & 59.5 & 64.9 & 68.4  & 68.2 \\
     SWDE Movie (HTML)      & 44.3 & 46.3 & 56.6  & 56.8 \\
     SWDE University (HTML) & 42.7 & 43.5 & 57.3  & 59.0 \\ 
     \hline
     \textbf{Average}  & \textbf{56.2} & \textbf{57.7} & \textbf{64.8} & \textbf{66.7} \\ 
     \bottomrule
    \end{tabular}}
\label{tab:ws_agg_ablation}
\vspace{-3mm}
\end{table}

%% file: sections/evaluations.tex
\subsubsection{Understanding the Tradeoff Space across Varied Language Models}
\label{sec:compare-models}
The are an increasing number of LLMs being made available. These models are trained by various providers each using distinct protocols \cite{liang2022helm}. To understand whether the tradeoffs we identified hold for different LLMs, we evaluate \systemname using three additional LLMs from three different providers: (1) \textbf{GPT-4} \cite{openai}, (2) \textbf{Anthropic Claude-V1} \cite{askell2021anthropic}, and (3)  \textbf{Jurassic Jumbo-2-Instruct} \cite{lieber2021jurassic}. Results are summarized in \cref{tab:across_models}.

\input{tables/across_models}

\textit{Overall results.}  
The quality with \verb|gpt-4| is comparable to that obtained using \verb|text-davinci-003|. Both the \systemnamea{} and \systemnamec{} quality decrease with \verb|claude| and \verb|jumbo|, consistent with the results of large-scale benchmarking efforts \cite{liang2022helm}, however the \textit{relative} quality of the two implementations are similar to \cref{tab:evaporate_vs_fm_openie}. Both appear to remain competitive in quality and the quality of the approaches appear to increase together.

We find the precision of \systemnamec{} remains high across models. \cref{alg:approach} helps \systemname{} filter the low quality functions and if this eliminates \text{all} the candidate functions, the attribute is excluded from the output table. We find that when an attribute \textit{is} included in the output, it has high precision, consistent with \cref{tab:evaporate_openie} where the precision with \verb|text-davinci-003| is almost 20 points higher than the recall. The average precision scores corresponding to \systemnamec{} in \cref{tab:across_models} are 70.9 (\verb|gpt-4|), 67.6 (\verb|claude|), and 50.9 (\verb|jumbo|) using \systemnamec{} and in contrast are 61.9 ( \verb|gpt-4|), 55.1 (\verb|claude|), and 49.9 (\verb|jumbo|) using \systemnamea{}, emphasizing a precision-recall tradeoff between approaches.

\textit{Understanding the errors.} Overall, \systemname{} relies on versatile reasoning capabilities (\textit{i.e.} to identify the schema and extract attribute values \textit{directly} from \textit{noisy} provided context, and the ability to synthesize code) and excitingly, the results validate that these capabilities co-exist within multiple model families. We investigate which of the required reasoning capabilities contributes to lower quality in comparison to \verb|text-davinci-003|. We find that the schema synthesis step plays a small role. Considering the top ranked schema attributes according to \systemname{}, we measure the average F1@$k$ between the predicted and gold sets of attributes, where $k$ is the number of gold attributes per setting. The average F1@$k$ for \verb|text-davinci-003| is 71.9, and the right-hand column of \cref{tab:across_models} shows the alternate models perform comparably.

We find the two main sources of errors are (1) the inability to generate a function for particular attributes, and (2) occasionally, low quality \textit{direct} extractions in particular cases (e.g., \verb|claude| may respond ``I'm not sure, please give me more information.'' in a ChatBot style, when prompted to extract an attribute value). The models we evaluate are optimized for ChatBot applications \cite{kocon2023chatgpt}.

%% file: tables/across_models.tex
\begin{table*}[t]
    \caption{OpenIE (Pair F1) results evaluating \systemname{} using alternate LMs from three model providers. For cost reasons, we apply \systemnamea{} to samples of 10 documents each. For fair comparison, we report the score of \systemnamec{} on the same sample instead of the full set of documents. $k$ is the number of gold attributes for the setting.}
    \vspace{-3mm}
    \centering
    \begin{tabular}{lccccc|ccccc|c}
     \toprule
    \multirow{2}{*}{Source (Format)}  &      \multicolumn{5}{c}{\centering \systemnamea{}} &
    \multicolumn{5}{c}{\centering \systemnamec{}} &
    \multicolumn{1}{c}{\centering \textsc{Schema ID}} \\
    &  \emph{FDA}  &  \emph{Wiki}  
    &  \emph{Movie}  &  \emph{University}  &  \emph{Enron}
    &  \emph{FDA}  &  \emph{Wiki}  
    &  \emph{Movie}  &  \emph{University}  &  \emph{Enron} & \emph{F1@$k$}\\
     \midrule

    % NOTE THESE SHOULD ALL BE ON THE 10 DOCUMENT SAMPLE

     OpenAI GPT-4 \cite{openai} & 
     59.2 & 40.5 & 35.1 & 56.1 & 92.7 & 
     57.5 & 61.4 & 54.9 & 57.2 & 85.5 & 67.3 \\
     
     Anthropic Claude-V1 \cite{askell2021anthropic} & 
     45.1 & 20.6 & 27.5 
     & 44.3 
     & 88.1 & 
       44.4 & 33.5 & 38.7 
     & 30.4  
     & 84.7 & 69.0 \\

     Jurassic Jumbo-2-Instruct \cite{lieber2021jurassic}  & 
     25.9 & 0.0 & 13.3 & 29.2 & 90.3 &
     1.2 & 0.0 & 20.6 & 18.6 & 85.7 & 
     62.3 \\

     \bottomrule
    \end{tabular} 
    %}
\vspace{-3mm}
\label{tab:across_models}
\end{table*}

%% file: sections/related_works.tex
\label{sec:related_work}
\paragraph{Structured Querying of Heterogeneous Data} 
Converting heterogeneous data to structured databases is a long standing data management problem \cite[inter alia.]{brin1998extraction, agichtein2000snowball, cafarella2007navigating}. 
In contrast to systems for knowledge base construction (KBC) or closed information extraction (IE)~\citep{yafooz2013managing, shin2015incremental}, which assume there is a predefined schema and focus on populating the database according to the schema, the setup we focus on relies on OpenIE.
OpenIE is the task of extracting useful facts without access to a predefined ontology (i.e. the types or categories of facts to be extracted) \cite{cohen2004ie, banko2007openie}. 
Given the breadth of input documents, the ability to construct a schema and populate the corresponding database on-the-fly is useful.

Existing systems for this problem introduce assumptions about the data-domain \cite{jayram2006avatar, derose2007dblife, deng2022domlm}, file-format (e.g., XML files)~\citep{garofalakis2000xtract}, or the syntactic patterns of useful facts \cite{brin1998extraction, agichtein2000snowball, etzioni2004knowitall, cafarella2007structured, mausam2012ollie, kolluru2020openie, niklaus2018openie, zhou2022survey}. For instance, in early systems, \citet{cafarella2007structured} focuses on facts expressed as triples (two entities with a descriptive string of their relationship in between) with hypernym ``is-a'' relationships between the entities. 
The recent deep learning based systems (1) require domain- and document-format specific training, (2) focus on reasoning over sentences, in contrast to long documents, and (3) rely on high quality linguistic tools (e.g. dependency parse, POS, NER) to help introduce structure over unstructured text \cite{kolluru2020openie, zhou2022survey}. 

For the narrower problem of generating structured views from \textit{web data} \cite{etzioni2004knowitall, cafarella2007structured, etzioni2008open}, the current state-of-the-art approaches use (1) distant supervision to train site-specific extraction models \cite{lockard2019ceres} (\textbf{domain specific}), and (2) rely on assumptions about where in the \textsc{HTML}-DOM attributes and values are located (\textbf{format specific}) \citet{lockard2020zeroshotceres, deng2022domlm}. We investigate the feasibility of a domain and document-format agnostic approach.

\vspace{-2mm}
\paragraph{Language Models for Data Management} Given the recency of in-context learning, there are few works exploring the benefits for data processing. Most closely related, \citet{chen2023symphony} presents a system for querying heterogeneous data lakes with in-context learning.  The proposed approach involves processing \textit{every document} with the LLM to extract values of interest. We propose an alternate approach and tradeoff space for processing data with LLMs.

Other recent work applies language models for tasks such as data wrangling~\cite{narayan2022can} or code generation for SQL queries~\cite{trummer2022codexdb}. Unlike \systemname{}, supporting various data formats, these prior approaches focus on relational data only and design manual prompts to demonstrate high quality.

\vspace{-2mm}
\paragraph{Data Programming} We build on work in data programming and weak supervision \cite{ratner2017snorkel}. \systemname{} automatically generates functions rather than using human-designed functions. We use WS on open ended tasks in contrast to the classification tasks considered in the prior work on automated WS \cite{varma2018snuba, boecking2021interactive}. We show how to use the LLM to handle abstensions and filter low quality functions.

%% file: sections/conclusion.tex
We propose \systemname, a system that uses LLM in-context learning to generate structured views of semi-structured data lakes. We identify and explore a cost-quality tradeoff between processing data directly with an LLM versus synthesizing and aggregating multiple code snippets for data processing. We present an algorithm and theoretical analysis for applying weak supervision to aggregate the code snippets. 
The code-based approach aims to exploit the structural redundancies that occur in corpora of semi-structured documents.
We validate \systemname on 16 unique data settings spanning 5 domains and 3 document formats, considering the cost, quality, and generality of the system. 
Our study highlights the promise of LLM-based data management systems.
\clearpage

%% file: sections/appendix.tex
\input{sections/appendix_details}

\input{sections/ws_appendix}

\input{sections/appendix_ablations}

\vspace{2mm}
\section{Additional Comparisons to Baselines} 

Here we provide additional quality and efficiency comparisons between \systemname and prior work.

\subsection{Additional Baselines from Prior Work} Here we study two additional baselines from prior work for OpenIE and ClosedIE respectively, beyond the Document-level IE methods discussed in Section \ref{sec:eval} (Table \ref{tab:evaporate_baselines}).
\label{sec:openie_baselines}
\paragraph{OpenIE} We apply the OpenIE6 system from \citet{kolluru2020openie}, which is a state-of-the art approach for OpenIE over unstructured (non-\textsc{HTML}) text. While this system is not designed for extracting structured views over heterogeneous data, we evaluate it qualitatively to understand how existing OpenIE systems perform in this setting.
\begin{enumerate}
    \item First, the system only handles well formatted sentences and struggles to extend to heterogeneous data types. It does not handle HTML documents and is difficult to apply to documents with complex formatting (like PDFs) where full sentences can be difficult to extract. Using SWDE College Prowler as an example, given the HTML input line \texttt{<td>Student Body Size:</td> <td class="stat"> <span id="..."> 6,504 </span> </td>}, OpenIE6 misses the student body size attribute and corresponding value.
    \item Second, even when documents contain full sentences, OpenIE6 extracts an extremely large set of relations for each document and does not prioritize attributes by relevance or enforce consistent attributes across documents. This makes it difficult to use the resulting relations to understand the contents of the documents or to do analysis. Using a sample FDA 510(k) document from our corpus as an example, OpenIE6 extracts 427 relations with 184 relations with confidence larger than 0.5. While some can be informative, a significant fraction of these relations are not useful for indexing and analysis across documents. For example, "(sample carryover; occurs; the results of the acceptor assay show interference)" is an extracted relation with confidence 0.99.
\end{enumerate}

\paragraph{ClosedIE} We study the effectiveness of span-extractor models, which are commonly used in QA systems to extract the information that is relevant to a user-query from a provided document context \cite{clark2020electra}. Given the ground truth attributes, we evaluate the ability of these models to extract their values from the relevant paragraphs. We evaluate the DebertaV3 Large model fine-tuned on the Squad 2.0 dataset, which achieves 90.8 F1 on the Squad 2.0 dev set in Table \ref{tab:deberta-qa-closedie}. We find text-davinci-003 (Table \ref{tab:fm_closedie_baseline}) and our \systemname{} function generation approach (Table \ref{tab:evaporate_openie}) significantly outperforms this pre-trained QA model on ClosedIE in all settings, over text and HTML documents.

\input{tables/qa_closedie}

\subsection{Efficiency Comparisons Between \systemname{} and Baselines}
\label{sec:efficiency_comparison}
\input{tables/efficiency-stats}

\edit{In this section, we compare the efficiency of \systemname and baselines like DOM-LM \cite{deng2022domlm}.
First, we breakdown the efficiency analysis in terms of the parameter count, pre-training cost, training costs, and inference costs. Then we use the cost break down to quanitatively compare \systemname to the baselines Table \ref{tab:efficiency-stats}. We also quantitatively compare \systemnamea and \systemnamec, highlighting how the decision hinges on the number of documents to be processed, as well as the number of attributes in the dataset. Finally, we compute estimated FLOP counts required to process documents from the five sources we considered (see Table \ref{tab:efficiency-data}). We use GPT3-175B as the language model in our calculations of costs for \systemname, since those are the primary results we report in the paper. We report experiments with multiple models in the paper and note that \systemname is flexible with respect to the underlying language model. This allows users to tradeoff quality and costs by using smaller language models (e.g. the smallest GPT model has with 175 million parameters) \cite{sizesOpenAI}.}

\edit{\begin{itemize}
    \item \textbf{Parameter count} The baseline uses a RoBERTa-Base LM backbone model, which is 125M parameters, while \systemname uses a 175B model.\footnote{While OpenAI has not released models, the \texttt{text-davinci-003} model is estimated to be 175B parameters \cite{sizesOpenAI}.} 
    We note that the baselines require training a new model for each new data domain, so the final parameter count for the baselines is $(\text{Number of Domains} \times \text{Model Size})$.
    \item \textbf{Pre-training costs} The LLMs used in \systemname see fewer tokens (300 billion tokens) \cite{brown2020language, ouyang2022instruct} than RoBERTa-base (2 trillion tokens) during pretraining \cite{roberta}. \citet{brown2020language} compares the pretraining of both models. Because of the disparity in parameter counts discussed above, the pre-training phase of the models used in \systemname{} is roughly $200$-times more FLOP-intensive than the baseline. However, users of \systemname or the baselines are not responsible for pre-training. Pre-trained models are commonly available via model hubs (\textit{e.g.} HuggingFace) and APIs, so pre-trained models can be reused in a wide variety of applications and by many users, amortizing the cost.
    \item \textbf{Training Costs} \systemname does not require any task-specific training while the baselines require that users train a new model for each data domain. Unlike pretraining, the task-specific training of the baselines would likely need to be performed on users' local compute. The exact cost of training is hard to estimate as it depends on the number of training epochs and the amount of training data. Additionally, collecting the requisite training data is a manual and time-intensive process. Specifically, SWDE Movies has data from 8 websites and SWDE Universities has data from 5 websites. Under the ``zero-shot'' protocol specified in \citet{deng2022domlm}, DOM-LM trains on 10\% of the data from 6 and 3 websites respectively for Movies and Universities, and evaluates on the heldout websites within each domain. Nonetheless, DOM-LM underperforms \systemname{}, which uses no task-specific labeled training data.
    \item \textbf{Inference costs} Inference FLOP requirements grow linearly in the model parameter count, so the cost of running inference on a single document with an LLM is significantly more expensive than with a baseline model. As shown in Table \ref{tab:efficiency-stats}, both DOM-LM and \systemnamea require $O(n)$ inference calls to process $n$ documents. On the other hand, \systemnamec only requires $O(m)$ inference calls in order to generate extraction functions which can then be applied to the entire corpus of documents. Thus, as the the number of documents to be processed grows, the inference cost of baseline methods eventually surpass the inference costs of \systemnamea.
\end{itemize}}

\edit{\paragraph{Quantifying the costs}
First we note that the FLOP requirement for pretraining RoBERTa and GPT-3 are reported in \cite{brown2020language}: 
\begin{itemize}
    \item \textbf{RoBERTa} The total training FLOPS is 1.50E+21. The inference FLOPS per token is $2 \times \text{Number of Parameters}$, which is $0.250$ GFLOPS. 
    \item \textbf{GPT3-175B} The total training FLOPS is 3.14E+23. The inference FLOPS per token is $2 \times \text{Number of Parameters}$, which is $350$ GFLOPS.
\end{itemize}}

\edit{The cost of applying \systemname and baselines to a document corpus varies depending on the number of documents $n$, the number of attributes $m$, and the model architecture used. The costs are summarized in Table \ref{tab:efficiency-stats}.}
\edit{\begin{itemize}
    \item \textbf{DOM-LM} The inference cost is computed by $$n \times \frac{\text{tokens}}{document} \times 0.250 \text{GFLOP}$$ where 0.250GFLOP is the inference cost per token of DOM-LM.
    \item \textbf{\systemnamec} The inference cost is computed by $$m \times P \times \frac{\text{tokens}}{chunk} \times 350 \text{GFLOP}$$ 
    where $P$ is the number of inference calls required per attribute and 350 GFLOP is the inference cost per token of \systemname, see Table \ref{tab:efficiency-stats}. Note that for our evaluated implementation $P \approx 10c$ for a small number $c$. 
    To understand where $c$ comes from, recall $D_{eval}$ is 10 documents. We generate candidate functions on chunks of each document and directly apply the LLM to each document so that we can score the function outputs against the LLM outputs.). 
\end{itemize}}
\edit{The inference FLOP counts for DOM-LM and \systemnamec fall in the same order of magnitude, though which method is more costly depends on the document source and the number of attributes. Users can select the most efficient method for their setting using the above cost framework. }

\vspace{3mm}
\section{Extensions to \systemname}

\edit{In this section, we provide experiments for three extensions to the \systemname system including 1) operating over a dataset that contains a mixture of the document sources (e.g. a mixture of SWDE Movies websites) (Section \ref{sec:source-clustering}), 2) further cleaning the output extractions from \systemname (for instance by converting them into atomic attributes) (Section \ref{sec:schema_cleaning_extensions}) and 3), using domain specific prompts to further control the system quality (Section \ref{sec:domain-specific}). and a summarizing discussion of future directions \ref{sec:discussion}. }

\subsection{Recovering Document Sources}
\label{sec:source-clustering}
\edit{In this section, we demonstrate how we can recover the sources from an unlabeled mixture of documents.}

\edit{In our experiments, we process each source with \systemname separately, instead of mixing them together. This more closely matches the experimental setup of the baselines like DOM-LM~\cite{deng2022domlm}. However, in practice, we may not have access to the sources of the documents in our data lake. To verify that we would be able to recover document sources from an unlabeled data lake, we explore whether we can use unsupervised methods to group documents by source.}

\edit{First, we mix the documents from the websites of the Movie and University subsets of SWDE. This provides an unlabeled ``data lake" with distinct document sources. Note that many of the sources come from the same domain, towards increasing the difficulty of distinguishing the sources. We apply a standard TF-IDF vectorizer and K-means clustering to the mixture of documents. Finally, we measure the adjusted rand index (ARI), a standard clustering metric which allows for random permutations in the cluster labels. We achieve a perfect ARI of 1.0, meaning we were able to exactly recover the document sources without any labeled data or supervision.}

\edit{In Figure \ref{fig:clusters}, we include a T-SNE visualization of the document TF-IDF vectors. This plot illustrates the separation in representation space between documents from different websites. }

\vspace{2mm}
\subsection{Atomic Schema Cleaning Extensions}
\label{sec:schema_cleaning_extensions}
One further extension of schema generation is atomic schema cleaning. \systemname{} generates a set of candidate attributes which, in some cases, are complex and can be decomposed into cleaned, atomic attributes. For example, an extracted attribute of ``born'' from the Wiki NBA setting, has the following form: <birth\_date> (age) <location>. Decomposing the ``born'' attribute into three separate attributes (e.g., birth date, age and birth location) would enable users to ask queries such as --- How many players in the NBA were born in Ohio? --- that would otherwise be unanswerable with the existing schema. As such, decomposing the complex attributes into cleaned, atomic attributes increases the utility of the resulting schema and extractions for analysis and indexing. Prior work~\citep{narayan2022can} has demonstrated that FMs can be useful for data transformation task. Schema decomposition can be viewed as an instantiation of data transformation, suggesting that such an operation could be completed using a FM.

We manually clean the ground truth complex attributes and values in the Wiki NBA setting and construct the ground truth atomic attribute and values. We find that after cleaning there are 35 atomic attributes for Wiki NBA, decomposed from the 19 complex attributes.

For our method, we prompt the expensive FM (in this case text-davinci-003 from OpenAI) to decompose the complex attribute and values into a list of atomic attributes and values, for a single example of each complex attribute and value. To save computation cost, we then use the large LLM schema cleaning result from one example to prompt a smaller, less expensive FM (in this case the text-curie-001 model from OpenAI) to extract the cleaned values for the remainder of documents. We provide the smaller, less expensive FM with the complex attribute, the cleaned attribute to extract, and a one-shot example from the expensive FM.

To measure the performance of schema cleaning, we construct pairs of \verb|(file, value)| for all files and values and compute the precision, recall, and F1 as in the OpenIE setting against the ground truth atomic attributes and values. We do not include the \verb|attribute| in the relations to score, because the extracted values are generally unique and we want to avoid penalizing generated atomic attribute names that differ from the ground truth but are still correct. As a baseline before our atomic schema cleaning step, we score the ground truth complex values against the ground truth atomic values and find it achieves an F1 of 21.0, since many values are not atomic. Applying our atomic schema cleaning methodology to the ground truth complex values decomposes them into atomic attributes, qualitatively improving the usability of the attributes. The resulting predicted atomic attributes achieve an F1 of 57.5 when scored against the ground truth atomic values. 

\subsection{Improving Quality via Domain Specific Prompts}
\label{sec:domain-specific}

\edit{A potential direction to improve the quality of \systemname is to replace the fixed, generic prompts with domain (e.g. Movies) or dataset (e.g. Rotten Tomatoes website) prompts. }

\edit{Towards demonstrating this opportunity, we update the Schema Identification prompt (Prompt \ref{sec:prompt-1}) to include in-context examples that discuss the Movies domain. The resulting prompt is provided in Section \ref{sec:prompt-dom}. We apply the single prompt to all 8 SWDE Movies websites (i.e. the prompts are not customized on a per-website level, but rather at the Movies domain-level). We observe this results in 4.5 Recall@K (8\%) improvement, averaged across the 8 settings. Results by website are provided in Table \ref{tab:domain_specific}. }

\input{tables/domain_specific}

\input{sections/using}

\section{Example Generated Candidate Functions}

\subsection{Case Study of Function Generation and Aggregation}

\label{sec:agg_case_study}

\edit{Here we provide two end-to-end examples of 1) generated candidate functions, 2) the outputs collected from different candidate functions and fed into the weak supervision algorithm, and 3) the resulting outputs. The first is an example attribute that achieves high quality under \systemname and the second is an attribute that achieves low quality.}

\edit{\textbf{FDA Setting} First we consider the 510k attribute in the FDA reports dataset.} 

\edit{Five of the ten generated functions are shown below:}
\begin{mdframed}
\begin{codeframe}
\begin{lstlisting}
def get_510_k_number_field(text: str):
    """
    Function to extract 510(k) number. 
    """
    pattern = r"K\d+"
    return re.findall(pattern, text)

def get_510_k_number_field(text: str):
    """
    Function to extract the 510(k) number.
    """
    pattern = r'510\(k\) Number:\s*(K\d+)'
    return re.findall(pattern, text)[0]

def get_510_k_number_field(text: str):
    """
    Function to extract 510(k) number. 
    """
    pattern = r'(?:510\(k\))\s*number\(s\):\s*(.*)'
    result = re.findall(pattern, text)
    return result

def get_510_k_number_field(text: str):
    """
    Function to extract the 510(k) number.
    """
    pattern = r'K\d{6}'
    return re.findall(pattern, text)

def get_510_k_number_field(text: str):
    """
    Function to extract the 510(k) number.
    """
    pattern = r'510\(k\) Number:\s*(k\d+)'
    return re.findall(pattern, text)[0]
\end{lstlisting}
\end{codeframe}
\end{mdframed}

\vspace{2mm}
\edit{Five examples of relevant snippets from FDA reports in the dataset are shown below as examples for the case study:}
\begin{mdframed}
\begin{codeframe}
\begin{lstlisting}

Report: K143467.txt
Content snippet:
ASSAY AND INSTRUMENT COMBINATION TEMPLATE 
A. 510(k) Number: 
k143467 

Report: K171742.txt
Content snippet:
DECISION MEMORANDUM 
 
A. 510(k) Number: 
 
K171742 

Report: K170491.txt
Content snippet:
ASSAY AND INSTRUMENT COMBINATION TEMPLATE 
A. 510(k) Number: 
K170491 

Report: K171641.txt
Content snippet:
DECISION SUMMARY 
A. 510(k) Number: 
K171641 

Report: K162526.txt
Content snippet:
ASSAY ONLY TEMPLATE 
A. 510(k) Number: 
k162526 

\end{lstlisting}
\end{codeframe}
\end{mdframed}

\vspace{2mm}
\edit{The resulting outputs of the functions applied to the reports are shown below. Note how 6 of 10 functions fail to extract the 510k number for report K143467.txt -- simply taking the majority vote across the functions would yield  an empty-value in the \systemname output. Further, how one function provides an incorrect prediction ``K2'' on report K162526.txt. The goal of the weak supervision algorithm is to model the function predictions across the dataset to determine which to rely on for different documents.}
\vspace{2mm}
\begin{mdframed}
\begin{codeframe}
\begin{lstlisting}

Report: K143467.txt
Predictions: ['', '', '', '', '', '', 'k143467', 'k143467', 'k143467', 'k143467']

Report: K171742.txt
Predictions: ['K171742', 'K171742', 'K171742', 'K171742', 'K171742', 'K171742', '', '', '', '']

Report: K170491.txt
Predictions: ['K170491', 'K170491', 'K170491', 'K170491', 'K170491', 'K170491', '', '', '', '']

Report: K171641.txt
Predictions: ['K171641', 'K171641', 'K171641', 'K171641', 'K171641', 'K171641', '', '', '', '']

Report: K162526.txt
Predictions: ['', '', '', '', '', 'K2', 'k162526', 'k162526', 'k162526', 'k162526']

\end{lstlisting}
\end{codeframe}
\end{mdframed}

\vspace{2mm}
\edit{We then apply the weak supervision algorithm (Algorithm \ref{alg:approach}) to the matrix of function outputs across all documents in the dataset (i.e. a matrix of shape $100 \times 10$ for the $100$ documents and $10$ candidate functions for the FDA setting). Because each function can output widely different values and the values differ across documents, the set of unique predictions per document varies. However, the label model expects a 1) fixed number of 2) constant (i.e. same for every document) classes. We handle the expectation for constant classes by \textit{symmetrizing} the label model as described in Section \ref{sec:approach-codeplus}. We handle the expectation for a fixed number of classes by taking the most frequently occurring $k$ (e.g. $k=5$) classes per document. If there are $<k$ unique predictions for a documenet, we fill the input vector with dummy values. For instance, this results in the following, given the $10$ outputs per document shown in the prior step:}

\begin{mdframed}
\begin{codeframe}
\begin{lstlisting}

Report: K143467.txt
Inputs: ['', 'k143467', 'dummy-1', 'dummy-2', 'dummy-3']

Report: K171742.txt
Inputs: ['K171742', '', 'dummy-1', 'dummy-2', 'dummy-3']

Report: K170491.txt
Inputs: ['K170491', '', 'dummy-1', 'dummy-2', 'dummy-3']

Report: K171641.txt
Inputs: ['K171641', '', 'dummy-1', 'dummy-2', 'dummy-3']

Report: K162526.txt
Inputs: ['', 'K2', 'k162526', 'dummy-1', 'dummy-2']

\end{lstlisting}
\end{codeframe}
\end{mdframed}

\vspace{2mm}
\edit{Finally, we learn the label model on the unlabeled inputs as shown above. The label model results in the following predictions on the five samples. The predictions are returned to the user in the output of \systemname.}
\vspace{2mm}

\begin{mdframed}
\begin{codeframe}
\begin{lstlisting}
Report: K143467.txt
Output: k143467

Report: K171742.txt
Output: K171742

Report: K170491.txt
Output: K170491

Report: K171641.txt
Output: K171641

Report: K162526.txt
Output: k162526

\end{lstlisting}
\end{codeframe}
\end{mdframed}
\vspace{2mm}
\edit{\textbf{Wikipedia Setting} Next we provide the same demonstrations for the ``born'' attribute in the NBA Players Wikipedia dataset.}

\edit{We provide examples of generated functions below:}
\vspace{2mm}

\begin{mdframed}
\begin{codeframe}
\begin{lstlisting}
def get_born_field(text: str):
    """
    Function to extract born. 
    """
    pattern = r"Born(.*?)Nationality"
    result = re.findall(pattern, text, re.DOTALL)
    return result

def get_born_field(text: str):
    """
    Function to extract the born field.
    """
    soup = BeautifulSoup(text, parser="html.parser")
    born_field = soup.find('span', class_="bday")
    born_field = born_field.text
    return born_field

def get_born_field(text: str):
    """
    Function to extract born. 
    """
    pattern = r"born in (.*?)\."
    matches = re.findall(pattern, text)
    return matches
\end{lstlisting}
\end{codeframe}
\end{mdframed}

\vspace{2mm}

\edit{We provide relevant snippets of three unique player documents from the dataset as examples for the case study.}

\vspace{2mm}
\begin{mdframed}
\begin{codeframe}
\begin{lstlisting}
Document: Luis_Flores.html
Content snippet: <th class="infobox-label">Born</th><td class="infobox-data"><span style="display:none"> (<span class="bday">1981-04-11</span>) </span>April 11, 1981<span class="noprint ForceAgeToShow"> (age 42)</span><br/>

Document: Magic_Johnson.html
Content snippet: <th class="infobox-label" scope="row">Born</th><td class="infobox-data"><span style="display:none"> (<span class="bday">1959-08-14</span>) </span>August 14, 1959<span class="noprint ForceAgeToShow"> (age 63)</span><br/>

Document: Draymond_Green.html
Content snippet: <th class="infobox-label" scope="row">Born</th><td class="infobox-data"><span style="display:none"> (<span class="bday">1990-03-04</span>) </span>March 4, 1990<span class="noprint ForceAgeToShow"> (age 32)</span><br/>
\end{lstlisting}
\end{codeframe}
\end{mdframed}

\vspace{2mm}

\vspace{2mm}

\edit{The resulting outputs of the functions applied to the three unique player documents from the dataset are shown below. }
\vspace{2mm}
\begin{mdframed}
\begin{codeframe}
\begin{lstlisting}
Document: Luis_Flores.html
Predictions: ['April 11, 1981', 'April 11, 1981', 'April 11, 1981', 'April 11, 1981', 'April 11, 1981', 'April 11', 'April 11', '1981-04-11', '(1981-04-11) April 11, 1981 (age\xa041)San Pedro de Macoris, Dominican Republic', '(1981-04-11) April 11, 1981 (age\xa041)San Pedro de Macoris, Dominican Republic']

Document: Magic_Johnson.html
Predictions: ['August 14, 1959', 'August 14, 1959', 'August 14, 1959', '1959', '1959', 'August 14', '1959, August 14, in <a href="/wiki/Lansing, child did not have HIV, 1956, to Melissa Mitchell. Although Andre was raised by his mother', '1959-08-14', '(1959-08-14) August 14, 1959 (age\xa063)Lansing, Michigan, U.S.', '(1959-08-14) August 14, 1959 (age\xa063)Lansing, Michigan, U.S.']

Document: Draymond_Green.html
Predictions: ['March 4, 1990', 'March 4, 1990', 'March 4, 1990', '1990', '1990', 'March 4', '1990, March 4, with his then-girlfriend Jelissa Hardy.<sup class="reference" id="cite_ref-164"><a href="#cite_note-164">[164]</a></sup><sup class="reference" id="cite_ref-165"><a href="#cite_note-165">[165]</a></sup> In 2018, 2020.<sup class="reference" id="cite_ref-166"><a href="#cite_note-166">[166]</a></sup><sup class="reference" id="cite_ref-167"><a href="#cite_note-167">[167]</a></sup><sup class="reference" id="cite_ref-168"><a href="#cite_note-168">[168]</a></sup> They held their wedding ceremony on August 14', '1990-03-04', '(1990-03-04) March 4, 1990 (age\xa032)Saginaw, Michigan, U.S.', '(1990-03-04) March 4, 1990 (age\xa032)Saginaw, Michigan, U.S.']

\end{lstlisting}
\end{codeframe}
\end{mdframed}

\vspace{2mm}
\edit{For the examples above, the input vectors provided to the weak supervision model are as follows. Recall results the $10$ outputs per document (obtained from the $10$ candidate functions) are condensed to the top-$k$ (e.g. $k=5$) most frequently occurring function outputs. }

\begin{mdframed}
\begin{codeframe}
\begin{lstlisting}

Document: Luis_Flores.html
Input: ['April 11, 1981', 'April 11', '1981-04-11', '(1981-04-11) April 11, 1981 (age\xa041)San Pedro de Macoris, Dominican Republic']

Document: Magic_Johnson.html
Inputs: ['August 14, 1959', '1959', August 14, in <a href="/wiki/Lansing, child did not have HIV, 1956, to Melissa Mitchell. Although Andre was raised by his mother', '1959-08-14', '(1959-08-14) August 14, 1959 (age\xa063)Lansing, Michigan, U.S.']

Document: Draymond_Green.html
Inputs: ['March 4, 1990', '1990', 'with his then-girlfriend Jelissa Hardy.<sup class="reference" id="cite_ref-164"><a href="#cite_note-164">[164]</a></sup><sup class="reference" id="cite_ref-165"><a href="#cite_note-165">[165]</a></sup> In 2018, 2020.<sup class="reference" id="cite_ref-166"><a href="#cite_note-166">[166]</a></sup><sup class="reference" id="cite_ref-167"><a href="#cite_note-167">[167]</a></sup><sup class="reference" id="cite_ref-168"><a href="#cite_note-168">[168]</a></sup> They held their wedding ceremony on August 14', '1990-03-04', '(1990-03-04) March 4, 1990 (age\xa032)Saginaw, Michigan, U.S.', '(1990-03-04) March 4, 1990 (age\xa032)Saginaw, Michigan, U.S.']

\end{lstlisting}
\end{codeframe}
\end{mdframed}

\vspace{2mm}
\edit{Applying the learned label model to the function predictions, the outputs are returned to the user by \systemname are shown below. Note that this is an example of a difficult attribute with lower quality results.}
\vspace{2mm}

\begin{mdframed}
\begin{codeframe}
\begin{lstlisting}
Document: Luis_Flores.html
Output: 'april 11 1981'

Document: Magic_Johnson.html
Output: '1959 08 14 august 14 1959 age\xa063lansing michigan u.s.'

Document: Draymond_Green.html
Output: '1990 03 04 march 4 1990 age\xa032saginaw michigan u.s.'
\end{lstlisting}
\end{codeframe}
\end{mdframed}

Note that the cleaning steps discussed in Section \ref{sec:schema_cleaning_extensions}, could help convert these noisy outputs.

\subsection{Generated Functions}
\label{sec:example_functions}

\edit{Here we provide additional examples of the candidate functions generated by \systemname. Note that the displayed functions are randomly selected and not filtered for quality (some may fail or achieve low quality) to give a representative sampling.}

\vspace{2mm}
\edit{First, within the FDA setting, the following functions are generated for the ``intended use'' attribute.}
\vspace{2mm}

\begin{mdframed}
\begin{codeframe}
\begin{lstlisting}
def get_intended_use_field(text: str):
    """
    Function to extract intended use. 
    """
    intended_use_field = re.findall(r'Intended Use(.*?)\n\n', text, re.DOTALL)
    return intended_use_field

def get_intended_use_field(text: str):
    """
    Function to extract intended use. 
    """
    intended_use_field = re.findall(r'Intended Use\s*(.*?)\s*Similar', text, re.DOTALL)
    return intended_use_field

def get_intended_use_field(text: str):
    """
    Function to extract the intended use field.
    """
    parts = text.split("G. Intended Use:")[-1]
    intended_use_field = parts.split("F. Regulatory Information:")[0]
    return intended_use_field
\end{lstlisting}
\end{codeframe}
\end{mdframed}

\vspace{2mm}
\edit{Second, within the Wikipedia NBA setting, the following functions are generated for the ``college'' attribute.}
\vspace{2mm}

\begin{mdframed}
\begin{codeframe}
\begin{lstlisting}
def get_college_field(text: str):
    """
    Function to extract the college field.
    """
    soup = BeautifulSoup(text, parser="html.parser")
    college_field = soup.find('th', string="College").find_next_sibling('td').text
    return college_field

def get_college_field(text: str):
    """
    Function to extract college. 
    """
    pattern = r'<i>(.*?)</i>'
    college_field = re.findall(pattern, text)
    return college_field

def get_college_field(text: str):
    """
    Function to extract the college field.
    """
    pattern = r'College Basketball at Sports-Reference.com'
    college_field = re.findall(pattern, text)
    return college_field[0]

def get_college_field(text: str):
    """
    Function to extract college. 
    """
    pattern = r"College.*?>(.*?)<"
    college_field = re.findall(pattern, text)
    return college_field
\end{lstlisting}
\end{codeframe}
\end{mdframed}

%% file: sections/appendix_details.tex
\clearpage

\section{Experimental Details}
\label{sec:protocol}
\subsection{Evaluation Protocol}
Because the baselines we compare against on SWDE require training data, they perform website-wise cross validation (\textit{i.e.} train on some websites and evaluate on others). They do this for several combinations such that every website appears in the evaluation set. \systemname, in contrast, does not require any training data, so we do not perform cross validation. To ensure a fair comparison, we simply evaluate \systemname{} on all websites so that our method is evaluated on the same examples as the baselines.

\subsection{Metrics} We describe the metrics we use to evaluate OpenIE and ClosedIE performance of our system. 

\textbf{Pair F1} For OpenIE, we report Pair F1 scores. Pair F1 is the standard metric for OpenIE systems \cite{lockard2019ceres}. The metric constructs \verb|(subject, value, predicate)|. The \verb|subject| is the document-filename in our setting. The \verb|predicate| is the attribute and the \verb|value| is the attribute value. The F1 score computes the F1 score between the sets of gold and predicted tuples. This assigns credit for \textit{exact-matches} between the attribute names and values extracted by the system and the ground truth --- it assigns no partial credit. 

Note that because \systemname{} first identifies a list of attributes then sequentially generates functions and extracts the values, the user can ``stop'' execution at any number of attributes. The stopping point determines the number of tuples included in the prediction set. This is not a property of prior systems that extract ``all or no'' tuples \cite[inter alia.]{lockard2019ceres, lockard2020zeroshotceres, kolluru2020openie, deng2022domlm, zhou2022survey}. For fair comparison we report performance at the number of gold attributes contained in the benchmark --- note that this is generally \textit{not} the number of attributes that maximizes the \systemname{}'s Pair F1 score.

\textbf{Text F1} For ClosedIE, we report Text F1 scores. Text F1 is the standard metric for extractive tasks and we use the exact implementation released by \citet{rajpurkar2022squad}. The metric tokenizes the prediction and gold strings and computes a token-wise F1 score.

Recall we select the F1 at the number of gold attributes, rather than at the number that gives the highest score).

\section{Dataset Construction}
\label{sec:benchmarks}
Below we describe how each of the evaluation benchmarks is obtained. We also release the suite of benchmarks along with the system code.
\subsection{FDA}

For the FDA setting, we randomly sampled a dataset of 100 FDA 510(k) premarket notification submission PDFs for medical devices with substantially equivalent predicate devices since 1996 from the FDA website: \url{https://www.accessdata.fda.gov/scripts/cdrh/cfdocs/cfpmn/pmn.cfm}. We used the lightweight fitz library to convert this to text files. We asked 5 database graduate students to identify important attributes, defined as attributes useful for analysis that's present in at least a majority of documents. We collected the final set of 16 attributes as attributes agreed on by all graduate students. For these 16 attributes, we manually wrote functions to extract their value, then corrected errors by manual review. We defined the attribute value as the full content of information pertaining to that attribute, typically a value, sentence, or section.
    
\subsection{Wiki NBA}
For the Wiki NBA setting, we used the following SPARQL query over Wikidata to retrieve NBA articles. We then manually supplemented missing pages and filtered the results to only include pages about NBA players.

    \begin{lstlisting}
        # Q13393265 is for Basketball Teams
        # Q155223 is for NBA
        # P118 is league (https://www.wikidata.org/wiki/Property:P118) 
        SELECT ?item ?itemLabel ?linkcount WHERE {
            ?item wdt:P118 wd:Q155223 .
            ?item wikibase:sitelinks ?linkcount .
        FILTER (?linkcount >= 1) .
        SERVICE wikibase:label { bd:serviceParam wikibase:language "[AUTO_LANGUAGE],en" . }
        }
        GROUP BY ?item ?itemLabel ?linkcount
        ORDER BY DESC(?linkcount)
    \end{lstlisting}

We asked 5 database graduate students to identify important attributes, defined as attributes useful for analysis that's present in at least a majority of documents. We collected the final set of 19 attributes as the attributes agreed on by all graduate students. For these 19 attributes, we manually wrote functions to extract their value, then corrected errors by manual review. We defined the attribute value as the full content of information pertaining to that attribute, typically a value, sentence, or section. We use these as ground truth extractions in the main paper.

We noted that the resulting attributes were complex for multiple documents, so we included another set of ground truth. We asked a graduate student to write functions to parse compound attributes whose values mention multiple values (e.g. birth date and location under attribute "Born") into atomic attributes and values. We this ground truth to demonstrate an additional step of schema cleaning in Section \ref{sec:schema_cleaning_extensions}.

\subsection{Enron}

We download the Enron corpus from \url{http://www.cs.cmu.edu/~enron/} and apply no further processing. We generate a benchmark using all metadata in the email headers by manually writing functions.

\subsection{SWDE} We download SWDE from \url{https://github.com/cdlockard/expanded_swde}. The benchmark includes the raw \textsc{HTML} from several websites with no further processing and ground-truth labels for selected attributes \cite{lockard2019ceres}. Because all the attributes are located within the root-elements of the \textsc{HTML} \textit{body}, excluding the information such as the \textsc{HTML} header, attributes described within tags (e.g. \verb|<a href='/year/2012/>'|, \verb|<title>|), and so on, we extend the original benchmark to include a more diverse set of attributes. We refer to the extended benchmark as SWDE Plus. 

%% file: sections/ws_appendix.tex
\section{Weak Supervision Details}
\label{sec:appendix_WS}

\paragraph{Objective} We detail the weak supervision (WS) algorithm used for aggregating the votes across generated functions. Let $\mathcal{D}$ be our \textit{unlabeled} dataset of documents from which we are extracting a particular attribute of interest. Let $y$ be a random variable representing the true attribute value. 
Let $\bm{\lambda}$ represent the outputs of our $m$ generated extraction functions $f \in \mathcal{E}$ on a particular document. 
Each $\lambda_i \in \bm{\lambda}$ is a function $\lambda_i: \mathcal{X} \rightarrow \mathcal{Y}$. Our goal is to use the vectors $\bm{\lambda}$, produced across documents $x \in \mathcal{D}$ to infer the true label $y$. Concretely, we seek, $\phi(x)$, a function that inputs $\bm{\lambda}$ and outputs a final prediction $\hat{y}$ for a document $x$.

With labeled data, i.e. where we had documents and their attribute values $\{(x_1, a_1), ..., (x_n, a_n)\}$, we could perform traditional supervised learn $\phi(x)$ that maps $\bm{\lambda}$ to $y$. However, in our unlabeled setting, the insight is to use the noisy estimates of $y$ produced by each of the functions to construct $\phi(x)$.

\paragraph{Standard WS Setup \cite{ratner2017snorkel}} WS models learn the latent variable graphical model on the distribution $\Pr(y, \{\lambda_1, ..., \lambda_m\} = \Pr(y, \bm{\lambda})$. In this model, $y$ is \textit{latent}, we cannot observe it. To produce $\hat{y}$, we concretely perform two steps:
\begin{itemize}
    \item \textbf{Learn the label model} We have access to $\bm{\lambda}$ and can use this to learn the \textit{label model} $P(\bm{\lambda}|y)$. 
    \item \textbf{Inference} We can then produce the estimate $\hat{y}$ by setting $\phi(x) = \argmax{y} \Pr(y|\bm{\lambda(x)}$
\end{itemize} 

First, we will discuss how to learn the label model. We parameterize $P(\bm{\lambda}|y)$ as follows: $$P(\bm{\lambda}|y) = \frac{1}{Z_{\theta}} \exp (\sum_{i=1}^{m} \theta d(\lambda_i, y))$$

 Intuitively when modeling our sources, we want to model how accurate a particular $\lambda_i$ is, when it makes a prediction. We are then going to want to weigh the predictions of the different sources proportional to their accuracies. 
 
$Z$ is a constant to normalize the probability distribution. The feature-vector $d$ can be broken down into:
\begin{itemize}
    \item $d_{\mathrm{lab}}$, representing how frequently $\lambda_i$ provides a prediction vs. abstains across documents. This \textbf{is} directly observable.
    \item $d_{\mathrm{corr}}$, which represents how frequently $\lambda_i$ and $\lambda_j$ yield the same prediction for the attribute value. This \textbf{is} directly observable.
    \item $d_{\mathrm{acc}}$, representing the accuracy of $\lambda_i$, or how frequently $\lambda_i$ agrees with $y$. Note that the accuracy is measured across documents for which the function provides a prediction and does not abstain. This \textbf{is not} directly observable. 
\end{itemize} 
$\theta$ are the parameters we aim to learn to combine the inputs in the feature vector. To learn this without access to $y$, we can minimize the negative
log marginal likelihood given the observed $\lambda_i$ outputs and solve with SGD, or use a closed-form solution \cite{fu2020fast}.

\subsection{Theoretical Analysis}
\label{sec:theory}

Here we analyze our proposed weak supervision algorithm (Algorithm \ref{alg:approach}) for \systemnamec{}.

\paragraph{Setup} Recall that in Algorithm \ref{alg:approach}, the LLM is 1) used to generate the candidate extraction functions $F = \{f_1, ..., f_m\}$, and 2) to directly extract values from a small set of $n$ documents, $D_{eval}$. Let $\{\hat{y}_1, ..., \hat{y}_n\}$ be the LLM's direct extractions over $D_{eval}$. The candidate functions are each applied to $x_i \in D_{eval}$ to produce $\{f_i(x_1), ..., f_i(x_n)\}$ for each $f_i \in F$.

The candidate function outputs $\{f_i(x_1), ..., f_i(x_n)\}$ are then scored against the LLM's direct extractions $\{\hat{y}_1, ..., \hat{y}_n\}$ to estimate the function accuracy: $$\hat{a} = \frac{1}{n}\sum_{j=1}^n \mathbf{1}(f_{i}(x_j) = \hat{y}_j)$$  

From \citet{varma2018snuba}, we can estimate whether the accuracies learned via weak supervision $\Tilde{a}$ will be close to the true function accuracies $a^*$. We can use the function accuracy estimated on $D_{eval}$ to estimate whether the accuracies $\Tilde{a}$ learned by the weak supervision model $\theta$ are close to the true function accuracies $a^*$, i.e. we can bound $||a^*-\Tilde{a}||_{\infty} < \gamma$. Concretely, the objective  is to provide a guarantee that $\gamma$ is the upper bound on the true error, given that the measured error $||\hat{a}-\Tilde{a}||_{\infty}$ is below some threshold $\epsilon$. Proposition 1 in \citet{varma2018snuba} states the following. Refer to \citet{varma2018snuba} for the proof.

\textit{We have $m$ functions with empirical accuracies $\hat{a}$ and learned accuracies $\Tilde{a}$, and the measured error is below some threshold $\epsilon$. Then, if each function labels a minimum of $$n \geq \frac{1}{2(\gamma-\epsilon)}\log(\frac{2m^2}{\delta})$$ datapoints, the weak supervision label model will succeed in learning accuracies such that $||a^* - \Tilde{a}||_{\infty} < \gamma$ with a probability $1 - \delta$.}

\paragraph{Extension} However, in our setting, $\hat{y}_i$ may not be equal to the true extraction $y_i^*$ for $d_i \in D_{eval}$ because the LLM is being used to score the candidate functions, rather than a gold-standard labeled dataset. Suppose the LLM introduces its own error rate $e$, then our objective is to modify the above statement to account for this source of error. Intuitively, as $e$ increases, we need more datapoints to obtain a better estimate of the accuracies.

\textit{Assumptions and Notation} We will assume the errors are randomly occurring across $x_i \in D_{eval}$ and that the data points in $D_{eval}$ are independent. Let $\hat{a}$ be the accuracy estimated using the noisy labels $\hat{y}$ on $D_{eval}$ and let $a^{\star}$ be the function accuracy given ground truth labels $y$ on $D_{eval}$.  We follow from \citet{varma2018snuba} below, but extend the theory to support our setting.

\paragraph{Proposition 1} \textit{We have $m$ functions with empirical accuracies $\hat{a}$, where the gold labels have error rate $e$, the learned accuracies are $\Tilde{a}$, and the measured error is below some threshold $\epsilon$. Then, if each function labels a minimum of $$n \geq \frac{1}{2(\gamma-\epsilon - e)}\log(\frac{2m}{\delta})$$ datapoints, the weak supervision label model will succeed in learning accuracies such that $||a^* - \Tilde{a}||_{\infty} < \gamma$ with a probability $1 - \delta$.}

\paragraph{Proof} We can decompose the true error $||a^* - \Tilde{a}||_{\infty}$ using the triangle inequality:

$$P(||a^* - \Tilde{a}||_{\infty} > \gamma) \leq 
P(||a^* - \hat{a}||_{\infty} + ||\hat{a} - \Tilde{a}||_{\infty} > \gamma)$$
$$\leq P(||a^* - \hat{a}||_{\infty} + \epsilon)$$

Recall that for candidate function $f_i$, where $n_i$ is the number of data points on which $f_i$ did not abstain: $$\hat{a}_i = \frac{1}{n_i} \sum_{j:f_i(x_j) \neq \emptyset} \mathbf{1}(f_i(x_j) = \hat{y}_j)$$
where $\hat{y}_j$ is the direct LLM extraction on document $x_j \in D_{eval}$,  $f_i(x_j)$ is the function output on $x_j$, and $n_i$ is the set of examples where the function did not abstain (i.e. $f_i(x_j) \neq \emptyset$). Suppose $P(\hat{y}_j \neq y_j^*) = e$. We again apply the triangle inequality to include $e$ in our decomposition of the true error. 

$$P(||a^* - \Tilde{a}||_{\infty} > \gamma) 
\leq P(||a^* - a^{\star}||_{\infty} + 
||a^{\star} - \hat{a}||_{\infty} + 
||\hat{a} - \Tilde{a}||_{\infty} 
> \gamma)$$
$$\leq P(||a^* - a^{\star}||_{\infty} + \epsilon + e)$$

Taking a union bound across the $m$ functions:
$$P(||a^* - \hat{a}||_{\infty}+ \epsilon +e > \gamma) \leq \sum_{i=1}^m P(||a_i^* - a^{\star}||_{\infty} + \epsilon + e > \gamma)$$ 

Next, $a^{\star}$ for function $f_i$ is defined, with respect to the ground truth labels $y$ on $D_{eval}$, where $n_i$ is the number of data points on which $f_i$ did not abstain:
$$a^{\star} = \frac{1}{n_i} \sum_{j:f_i(x_j) \neq \emptyset} \mathbf{1}(f_i(x_j) = y_j)$$ 

Substituting $a^{\star}$ into our bound:
$$P(||a^* - \hat{a}||_{\infty}+ \epsilon +e > \gamma) \leq \sum_{i=1}^m P(||a_i^* - a^{\star}||_{\infty} + \epsilon + e > \gamma)$$ 
$$\leq 
\sum_{i=1}^m P(|a_i^* - \frac{1}{n_i} \sum_{j:f_i(x_j) \neq \emptyset} \mathbf{1}(f_i(x_j) = y_j)| > \gamma - + \epsilon - e)$$ 

Next we can apply Hoeffding's inequality:
$$\leq \sum_{i=1}^m 2\exp(-2(\gamma - \epsilon - e)^2n_i)$$ 
$$\leq 2m\exp(-2(\gamma - \epsilon - e)^2\min(n_1, ...n_m))$$ 

Finally, rearranging the terms to isolate $n$, where $n = \min(n_1, ...n_m)$:
$$P(||a^* - \Tilde{a}||_{\infty} > \gamma) \leq 2m\exp(-2(\gamma - \epsilon - e)^2n$$
$$\delta \leq 2m\exp(-2(\gamma - \epsilon - e)^2n)$$

Taking the $\log$ of both sides each side is $< 1$ in value:
$$\log(\delta) \geq \log(2m\exp(-2(\gamma - \epsilon - e)^2n))$$
$$\log(\delta) \geq \log(2m) + \log(\exp(-2(\gamma - \epsilon - e)^2n)$$

Negating both sides:
$$\log(2m) - \log(\delta) \leq 2(\gamma - \epsilon - e)^2n$$
$$\frac{1}{2(\gamma - \epsilon - e)^2}\log(\frac{2m}{\delta}) \leq n$$

\paragraph{Analysis} As the number of aggregated candidate functions $m$ increases, then the number of points each function should label for learning the label model, i.e. the size of $D_{eval}$, should be increased. As the underlying LLM generating $\hat{y}$ increases in accuracy, we can decrease the size of $D_{eval}$.

%% file: sections/appendix_ablations.tex
\vspace{2mm}
\section{Additional Ablations for \systemname{}} 
\subsection{Validating the Quality of FM $\mathcal{F}$}
\label{sec:upperbound}
\input{tables/fm_closedie_baseline}

Because our approach for scoring the generated functions relies on comparing to the extractions produced by directly processing the document with the FM $\mathcal{F}$ (Section \ref{sec:approach-direct}), in Table \ref{tab:fm_closedie_baseline}, we evaluate the quality of $\mathcal{F}$. In contrast to the ``End-to-End'' prompt baseline (Prompt  \ref{sec:prompt-1}), in this experiment, the FM is prompted with a specific attribute to extract. This prompt is shown in Appendix \ref{sec:microprompts}.

Next, we compare the quality that is achieve using Algorithm \ref{alg:approach} by using \textit{ground truth labels} on $D_{eval}$ (containing 10 documents) to \systemname{} achieves using FM $\mathcal{F}$. Recall that in Algorithm \ref{alg:approach}, the ``labels'' from  $\mathcal{F}$ are used to both estimate the fraction $e$ and the scores for functions $f \in F$. We find that the upper bound quality, using ground-truth labels, is 5.1 F1 points higher as shown in Table \ref{tab:gold_upper_bound}.

\input{tables/gold_comparison}

\subsection{Varying Number of Candidate Functions}
\label{sec:varying-ncandidates}
\edit{
In this section, we explore how varying the number of candidate functions generated by \systemnamec affects the quality and efficiency of the system. }

\begin{figure}
    \centering
\includegraphics[width=0.7\columnwidth]{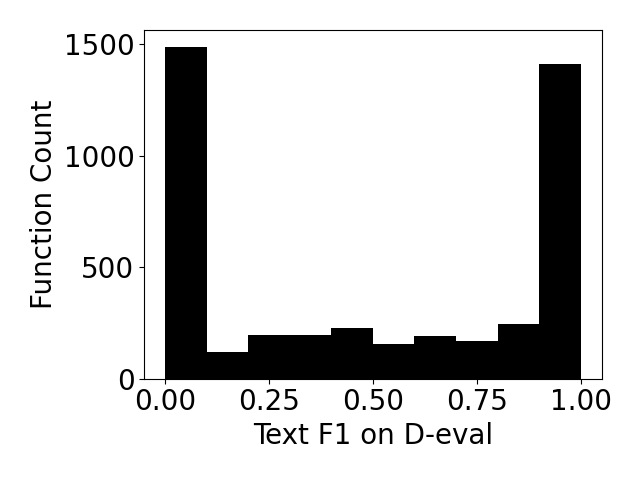}
    \vspace{-4mm}
    \caption[width=\columnwidth]{\edit{Across the gold attributes and settings, we report the estimated quality of the generated extraction functions.}}
    \label{fig:best_extractor_split}
    \vspace{-3mm}
\end{figure}

\edit{Recall from \ref{sec:approach_aggregation}, that \systemnamec generates a diverse set of candidate extraction functions a diverse set of \textit{function candidates} $F = \{f_1, f_2, ... f_{k}\}$, with variable quality (Figure \ref{fig:best_extractor_split}). The outputs of the generated functions are then aggregated using a weak supervision algorithm (Algorithm \ref{alg:approach}).} 

\input{tables/varying-ncandidates}
\input{tables/varying_nfunctions}

\edit{In our main experiments, we generate the pool of candidate functions and score the candidate functions on $d=10$ documents. We then feed the top $k=10$ highest scoring candidate functions to the weak supervision algorithm.}

\edit{First, we measure the quality of \systemnamec{} as we vary the number of documents $d$ used to generate and score the candidate functions. We evaluate for $d \in \{1, 3, 5, 10\}$. The results are reported in Table \ref{tab:vary_ncandidates}.}

\edit{Next, we measure the quality of \systemnamec{} as we vary the number of functions $k$ passed to the weak supervision algorithm. We evaluate for $k \in \{1, 3, 5, 10\}$. The results are reported in Table \ref{tab:vary_nfunctions}.}

\edit{The quality increases with both the number of seed documents used to generate and evaluate candidate functions, and with the number of candidate functions passed to the weak supervision algorithm.}

%% file: tables/fm_closedie_baseline.tex
\begin{table}[t]
    \caption{Quality and cost achieved through prompting OpenAI's Text-Davinci-003 model to extract specific, pre-defined attributes.}
    \centering
    \resizebox{0.48\textwidth}{!}{\renewcommand{\arraystretch}{0.98}
    \begin{tabular}{lcc|cc}
     \toprule
       \multirow{2}{*}{Source (Format)}  &  
       \multicolumn{2}{p{2.5cm}}{\centering \textsc{Quality}} & \multicolumn{2}{p{3cm}}{\centering \textsc{Cost / 10k Documents}} \\   &  \# Attributes  &\emph{F1}  & \emph{Tokens (M)} & \emph{Dollars (\$)}\\
    \midrule
     Enron Emails (\textsc{TXT})     & 15 & 85.3 & 140 & 2,790 \\
     FDA (\textsc{TXT})              & 16 & 78.0 & 241 & 4,816 \\
     Wiki NBA (\textsc{HTML})        & 19 & 84.6 & 328 & 6,559 \\
     SWDE Movies (\textsc{HTML})     & 25 & 84.4 & 359 & 7,174 \\
     SWDE University (\textsc{HTML}) & 33 & 72.6 & 379 & 7,586 \\ 
     \hline
     \textbf{Average} & \textbf{21.6}  & \textbf{79.9} & \textbf{289} & \textbf{5,785} \\ 
     \bottomrule
    \end{tabular}}
\label{tab:fm_closedie_baseline}
\end{table}

%% file: tables/gold_comparison.tex
\begin{table}[t]
    \caption{The left column includes our main results from Table \ref{tab:evaporate_openie} for reference. On the right, we evaluate the quality of \systemname{} when ground-truth labels are used instead to score functions and estimate the abstension fraction $e$ in Algorithm \ref{alg:approach}.}
    \centering
    \resizebox{0.48\textwidth}{!}{\renewcommand{\arraystretch}{0.98}
    \begin{tabular}{lcc}
     \toprule
       \multirow{1}{*}{Source (Format)}  &      \multicolumn{1}{p{1.4cm}}{\centering{OpenIE with $\mathcal{F}$}} &
       \multicolumn{1}{p{2.8cm}}{\centering{OpenIE with Ground-Truth}} \\
    \midrule
     Enron Emails (\textsc{TXT})      & 89.9 &  87.0 \\
     FDA (\textsc{TXT})               & 62.8 &  61.9 \\
     Wiki NBA (\textsc{HTML})         & 68.2  &  79.7 \\
     SWDE Movie (\textsc{HTML})       & 56.8 &  64.0 \\
     SWDE University (\textsc{HTML})  & 59.0 &  66.3 \\ 
     \hline
     \textbf{Average}  & \textbf{66.7} & \textbf{71.8}  \\ 
     \bottomrule
    \end{tabular}}
\label{tab:gold_upper_bound}
\end{table}

%% file: tables/varying-ncandidates.tex
\begin{table}[t]
    \caption{\edit{Quality using varying numbers of documents from which candidate functions are generated and scored.}}
    \small
    \centering
    \resizebox{0.48\textwidth}{!}{\renewcommand{\arraystretch}{0.98}
    \edit{\begin{tabular}{lcccc}
     \toprule 
       \multirow{1}{*}{Source Dataset (Format)}  &   
       1 &
       3 &
       5 & 
       10 \\
    \midrule
     FDA (TXT)             & 40.0 & 56.5 &  67.1 & 62.8 \\
     Enron Emails (TXT)    & 45.5 & 54.1 &  67.2 & 86.9 \\
     Wiki NBA (HTML)       & 64.3 & 61.4 & 64.7  & 68.2 \\
     SWDE Movies (HTML)    & 45.6 & 46.7 & 51.7  & 56.8 \\
     SWDE Universities (HTML) & 38.9 & 44.1 & 55.1  & 59.0 \\ 
     \hline
     \textbf{Average}  & \textbf{46.9} & \textbf{52.6} & \textbf{61.2} & \textbf{66.7} \\ 
     \bottomrule
    \end{tabular}}}
\label{tab:vary_ncandidates}
\end{table}

%% file: tables/varying_nfunctions.tex
\begin{table}[t]
    \caption{\edit{Quality achieved by passing varying numbers $k$ of the top-$k$ candidate functions to the weak supervision algorithm. For all runs, the pool of candidate functions are generated and scored by prompting on 10 seed documents.}}
    \small
    \centering
    \resizebox{0.48\textwidth}{!}{\renewcommand{\arraystretch}{0.98}
    \edit{\begin{tabular}{lcccc}
     \toprule 
       \multirow{1}{*}{Source Dataset (Format)}  &   
       1 &
       3 &
       5 & 
       10 \\
    \midrule
     FDA (TXT)             & 49.0 & 59.1 & 62.1 & 62.8 \\
     Enron Emails (TXT)    & 85.7 & 85.7 & 85.7 & 86.9 \\
     Wiki NBA (HTML)       & 69.9 & 67.2 & 68.9 & 68.2 \\
     SWDE Movies (HTML)    & 56.9 & 55.7 & 55.4 & 56.8 \\
     SWDE Universities (HTML) & 56.7 & 58.9 & 59.8 & 59.0 \\ 
     \hline
     \textbf{Average}  & \textbf{63.6} & \textbf{65.3} & \textbf{66.4} & \textbf{66.7} \\ 
     \bottomrule
    \end{tabular}}}
\label{tab:vary_nfunctions}
\end{table}

%% file: tables/qa_closedie.tex
\begin{table}[t]
\caption{Results as in Table \ref{tab:fm_closedie_baseline} using the DebertaV3 large model fine-tuned on the Squad2.0 dataset from HuggingFace.}
\centering
\resizebox{0.48\textwidth}{!}{\renewcommand{\arraystretch}{0.98}
\begin{tabular}{lcc}
\hline
Source (Format)        & \# Attributes & Closed IE F1 \\
\hline
Enron Emails (TXT)     & 15            & 53.7         \\
FDA (TXT)              & 17            & 56.5         \\
Wiki NBA (HTML)        & 19            & 50.2         \\
SWDE Movies (HTML)     & 30            & 43.5         \\
SWDE University (HTML) & 25            & 45.3         \\
\hline
\end{tabular}}
\label{tab:deberta-qa-closedie}
\end{table}

%% file: tables/efficiency-stats.tex
\begin{table*}[] 
\caption{\edit{Statistics on computational cost of the language models used by \systemname and baselines. The parameter counts and pre-training costs are sourced from \cite{brown2020language}.  \textit{Inference calls} is the number of LLM forward passes required to process $n$ documents with $m$ attributes. FLOP counts assume that DOM-LM is implemented with a RoBERTa backbone (as reported in the paper~\cite{deng2022domlm}) and that \systemname is implemented with text-davinci-003.}}
\edit{\begin{tabular}{llllll}
\hline
   & Parameter Count & Pre-training  Cost & Fine-tuning Cost  & Inference Calls & Inference Cost  \\ 
   & (B)                & (ZFLOP)          & (GFLOP per token) & (\# of Documents)  & (GFLOP per token) \\ \hline
DOM-LM~\cite{deng2022domlm}           & 0.125         &  1.50 & 11.250 & $O(n)$   & \multicolumn{1}{r}{0.250}     \\
\systemnamea & 175           & 341 & 0 & $O(n)$ & \multicolumn{1}{r}{350} \\
\systemnamec  & 175           & 341 & 0 & $O(m)$ & \multicolumn{1}{r}{350} \\ \hline
\end{tabular}}
\label{tab:efficiency-stats}
\end{table*}

\begin{table*}[]
\caption{\edit{Language model inference costs for processing 100K Documents with \systemname and baselines. Because document length and number of attributes, $m$, varies by source, we report costs for each of the sources included in our experiments. FLOP counts are based on the statistics reported in Table \ref{tab:efficiency-stats}. FLOP counts assume that DOM-LM is implemented with a RoBERTa backbone (as reported in the paper~\cite{deng2022domlm}) and that \systemname is implemented with text-davinci-003.}}
\edit{
\begin{tabular}{lrrrrrr}
\hline
                       & \multicolumn{6}{c}{Cost / 100K Documents}                                                                                                                                               \\
                       \\
Source (Format)        & \multicolumn{2}{c|}{DOM-LM~\cite{deng2022domlm}}                                 & \multicolumn{2}{c|}{\systemnamea}                        & \multicolumn{2}{c}{\systemnamec}                        \\
                       & \multicolumn{1}{l}{Tokens (M)} & \multicolumn{1}{l|}{PFLOP} & \multicolumn{1}{l}{Tokens(M)} & \multicolumn{1}{l|}{PFLOP}   & \multicolumn{1}{l}{Tokens (M)} & \multicolumn{1}{l}{PFLOP} \\ \hline
FDA (TXT)              & 145.6                          & \multicolumn{1}{r|}{364}   & 145.6                         & \multicolumn{1}{r|}{50,960}  & 1.9                            & 665                       \\
Enron Emails (TXT)     & 21.2                           & \multicolumn{1}{r|}{53}    & 21.2                          & \multicolumn{1}{r|}{7,420}   & 0.6                            & 210                       \\
WIki NBA (HTML)        & 650.1                          & \multicolumn{1}{r|}{1,625}  & 650.1                         & \multicolumn{1}{r|}{227,535} & 3                              & 1,005                     \\
SWDE Movie (HTML)      & 282.9                          & \multicolumn{1}{r|}{707}   & 282.9                         & \multicolumn{1}{r|}{99,015}  & 2.3                            & 805                       \\
SWDE University (HTML) & 190.1                          & \multicolumn{1}{r|}{475}   & 190.1                         & \multicolumn{1}{r|}{66,535}  & 1.9                            & 665                       \\ \hline
\end{tabular}}
\label{tab:efficiency-data}
\end{table*}

%% file: tables/domain_specific.tex
\begin{table}[t]
    \caption{\edit{Comparing Recall@$K$, where $K$ is the number of gold attributes for the dataset, on Schema Identification using generic, domain agnostic vs. domain specific in-context demonstrations for the 8 SWDE Movie websites.}}
    \small
    \centering
    \resizebox{0.48\textwidth}{!}{\renewcommand{\arraystretch}{0.98}
    \edit{\begin{tabular}{lccc}
     \toprule 
       \multirow{1}{*}{SWDE Movies Website}  &   
       Base &
       Domain &
       Difference \\
    \midrule
     Allmovie       & 62.7 & 65.2 & $+$ 2.5  \\
     AMCTV          & 88.2 & 83.9 & $-$ 4.3  \\
     Hollywood     & 14.4 & 7.9 & $-$ 6.5 \\
     iHeart Movies & 62.5 & 75.0 & $+$ 12.5 \\
     IMDB           & 70.4 & 71.6 & $+$ 1.2  \\
     Meta Critic    & 35.3 & 64.7 & $+$ 29.4 \\
     Rotten Tomatoes   & 66.7 & 58.3 & $-$ 8.4  \\
     Yahoo Movies         & 72.7 & 81.8 & $+$ 9.1  \\
     \hline
     \textbf{Average}  & \textbf{59.1} & \textbf{63.6} & \textbf{4.5}  \\ 
     \bottomrule
    \end{tabular}}}
\label{tab:domain_specific}
\vspace{-6mm}
\end{table}

%% file: sections/using.tex
\vspace{2mm}
\subsection{Discussion of Future Directions} 
\label{sec:discussion}

The goal of this study was to evaluate a simple, prototype system that uses LLMs to generate structured views from unstructured documents. We explored two fundamentally different implementation strategies, highlighting opportunities for future work in the space.

\textbf{New applications} Our findings demonstrate the promise of function synthesis as a way to mitigate cost when using LLMs. We study the problem of materializing a structured view of an unstructured dataset, but this insight may be applicable in a broader suite of data wrangling tasks. Many data wrangling tasks are high throughput applications, for which LLMs are not optimized. Future work should explore whether \textit{code synthesis} may enable general low cost solutions. 

\textbf{New function types} We dichotomized the \textit{direct extraction} and \textit{code synthesis} implementations. However, the lines between these approach may blur going forward. After all, the LLM could generate functions that invoke \textit{other models} --- for instance, the LLM may generate functions that call the \textsc{NLTK}, \textsc{Huggingface}, or even the \textsc{OpenAI} APIs. This naturally raises the question of how to characterize the cost of the generated functions, rather than assuming they are inexpensive. 

\textbf{Improving quality} Future work may consider iterative approaches to function generation. Concretely, when a generated function fails to compile or achieves low scores compared to the high quality LLM, we may be able to provide the compilation errors and/or high quality LLM responses in a prompt that encourages the LLM to generate an improved function. For instance, we may use Algorithm \ref{alg:approach} to score the quality of \textit{small} LLMs for performing the extractions. 

\edit{The goal of our work is to provide improved methods and infrastructure for using recent LLM technology. A complementary objective is to improve the quality of the LLM itself. The LLMs with high quality in-context learning results tend to be parameter intensive, which means that further training the model can be expensive. There are two main approaches in the literature that explore efficiently adapting the LLM itself to new data that we refer to: parameter-efficient fine-tuning and retrieval augmented generation. Parameter-efficient fine-tuning methods update a small number of model parameters, keeping the rest frozen, thereby significantly improving the efficiency of fine-tuning while still attaining strong results \cite{peft}. Retrieval-augmented generation methods retrieve relevant information and incorporate it into generation, thereby making it easier to update and change the information used by the LLM \cite{li2022rag}.}

%% file: sections/input_output.tex
\section{System Input and Output Diagrams}
\begin{figure*}
    \centering
    \includegraphics[width=\linewidth]{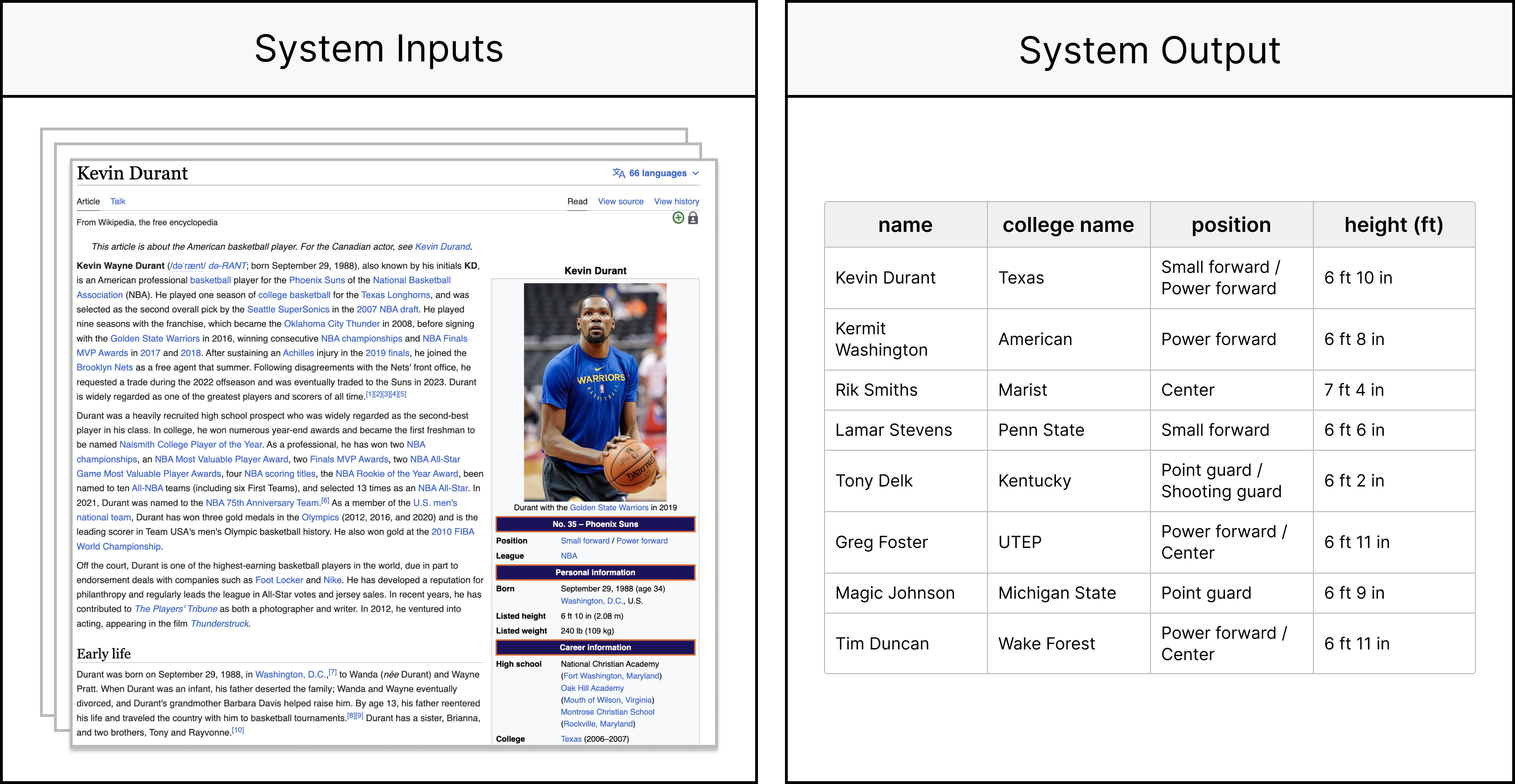}
    \caption[width=\linewidth]{Diagram depicting \systemname{} input and sample output on the Wikipedia NBA Players (HTML) setting.}
    \label{fig:input-output-wiki}
\end{figure*}

\begin{figure*}
    \centering
    \includegraphics[width=\linewidth]{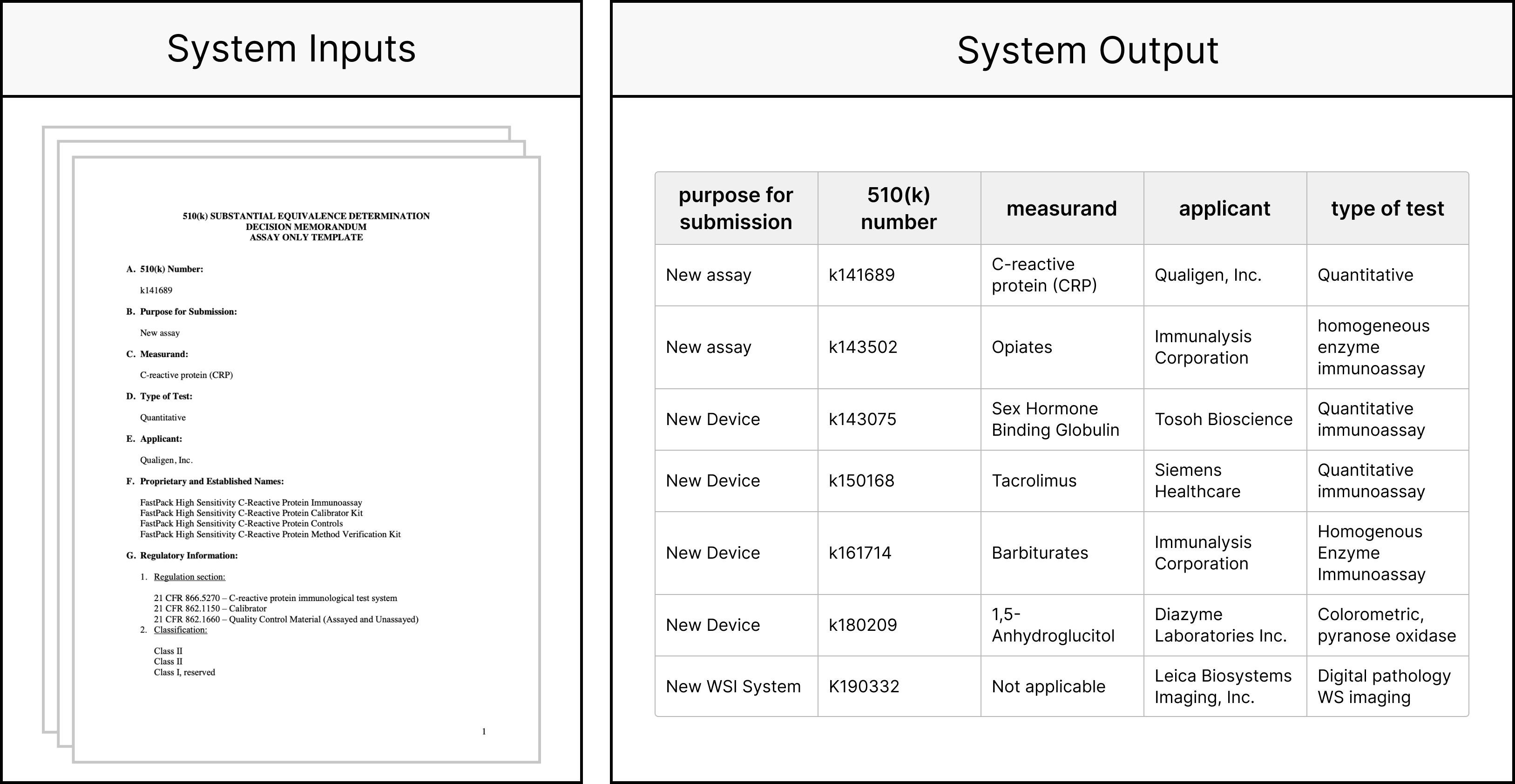}
    \caption[width=\linewidth]{Diagram depicting \systemname{} input and sample output on the Medical AI Device FDA Reports (TXT) setting.}
    \label{fig:input-output-fda}
\end{figure*}

\begin{figure*}
    \centering
    \includegraphics[width=\linewidth]{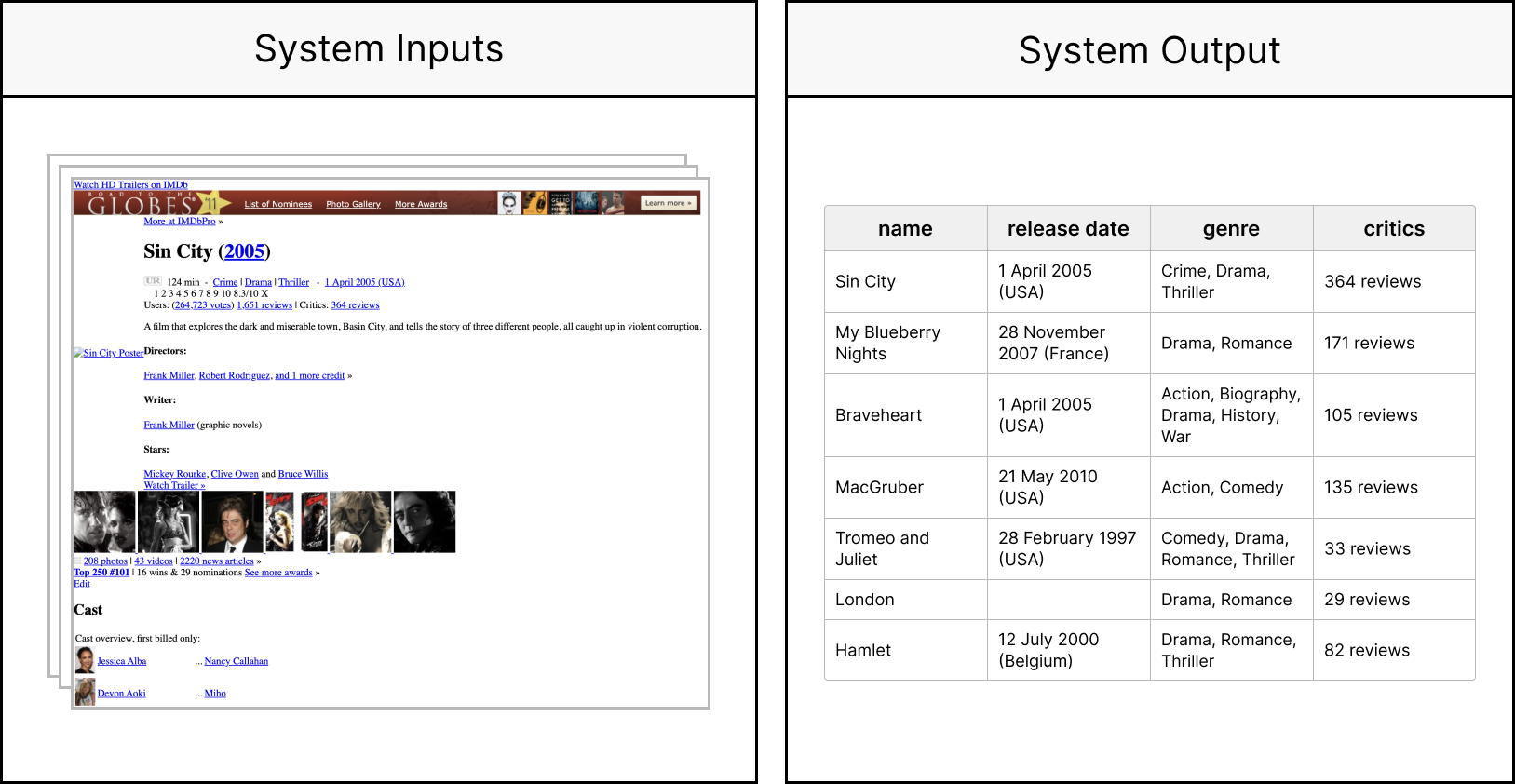}
    \caption[width=\linewidth]{Diagram depicting \systemname{} input and sample output on the SWDE Movies IMDB (HTML) setting.}
    \label{fig:input-output-imdb}
\end{figure*}

In Figures \ref{fig:input-output-wiki}, \ref{fig:input-output-fda}, and \ref{fig:input-output-imdb}, we include sample inputs and outputs for \systemname{}.

